\documentclass{article}

\usepackage{microtype}
\usepackage{graphicx}
\usepackage{booktabs} 
\usepackage{multirow}
\usepackage{amsmath}
\usepackage{caption}
\usepackage{subcaption}
\usepackage{adjustbox}

\usepackage{hyperref}

\usepackage[accepted]{mlsys2020}

\mlsystitlerunning{Benchmarking Deep Learning Inference Serving}

\begin{document}

\twocolumn[
\mlsystitle{InferBench: Understanding Deep Learning Inference Serving with an Automatic Benchmarking system}


\mlsyssetsymbol{equal}{*}

\begin{mlsysauthorlist}
\mlsysauthor{Huaizheng Zhang}{to}
\mlsysauthor{Yizheng Huang}{to}
\mlsysauthor{Yonggang Wen}{to}
\mlsysauthor{Jianxiong Yin}{nvidia}
\mlsysauthor{Kyle Guan}{bell}
\end{mlsysauthorlist}

\mlsysaffiliation{to}{Nanyang Technological University, Singapore}
\mlsysaffiliation{nvidia}{NVIDIA AI Tech Center}
\mlsysaffiliation{bell}{Nokia Bell Lab}

\mlsyscorrespondingauthor{Huaizheng Zhang}{huaizhen001@e.ntu.edu.sg}

\mlsyskeywords{Deep Learning, MLSys}

\vskip 0.3in

\begin{abstract}
Deep learning (DL) models have become core modules for many applications. However, deploying these models without careful performance benchmarking that considers both hardware and software's impact often leads to poor service and costly operational expenditure. To facilitate DL models' deployment, we implement an automatic and comprehensive benchmark system for DL developers. To accomplish benchmark-related tasks, the developers only need to prepare a configuration file consisting of a few lines of code. Our system, deployed to a leader server in DL clusters, will dispatch users' benchmark jobs to follower workers. Next, the corresponding requests, workload, and even models can be generated automatically by the system to conduct DL serving benchmarks. Finally, developers can leverage many analysis tools and models in our system to gain insights into the trade-offs of different system configurations. In addition, a two-tier scheduler is incorporated to avoid unnecessary interference and improve average job compilation time by up to 1.43x (equivalent of 30\% reduction). Our system design follows the best practice in DL clusters operations to expedite day-to-day DL service evaluation efforts by the developers. We conduct many benchmark experiments to provide in-depth and comprehensive evaluations. We believe these results are of great values as guidelines for DL service configuration and resource allocation.

\end{abstract}
]



\printAffiliationsAndNotice{}  

\section{Introduction}
\label{introduction}

Deep learning (DL) tools are transforming our life, as they are incorporated into more and more cloud services and applications \cite{tran2015learning, bahdanau2014neural, gatys2016image, silver2017mastering}. Facing this ever-increasing DL deployment demand, many companies, tech-giants or startups, are engaging in a fierce arm race to develop customized inference hardware \cite{jouppi2017datacenter}, software \cite{baylor2017tfx} and optimization tools \cite{chen2018tvm} to support DL services. Hence, there is a huge need to concurrently develop systems to study and benchmark their performance as guidelines for deploying high-performance and cost-effective DL services.


Unlike conventional web frameworks and techniques, deep learning (DL) techniques as well as their inference hardware and software platforms, are still evolving at a rapid pace. To address this challenge, an easy-to-use and highly deployable benchmark system is needed to assist DL developers in their daily performance evaluation tasks and service configuration. Also, the system must provide out-of-the-box methodologies to assess and understand the intricate interactions among DL models, hardware, and software across a variety of configurations (e.g., batch size and layer number) \cite{wang2020systematic}. The obtained results can be used as guidelines for future design.


Though existing benchmark studies have made substantial contributions to understand DL inference performance \cite{reddi2020mlperf, ignatov2019ai, coleman2017dawnbench}, they still do not adequately address the aforementioned challenges. MLPerf inference benchmark \cite{reddi2020mlperf}, as the current state-of-the-art solution, lacks the much-needed configurability, as it leaves the implementation details to developers. As a result, developers have to spend days or even weeks preparing a benchmark submission for a fair comparison. In addition, the results \cite{reddi2020mlperf} only showed the importance of hardware and software configuration but provided very limited insights and prescriptive guidelines for DL service configuration and resource allocation. Moreover, benchmarking performance using only one or two models optimized for specific inference tasks (e.g., a well-trained ResNet50 model \cite{he2016deep} for image classification) provides a very limited understanding of the impact of hyper-parameters (e.g., layer number and batch size) on the performance or resource utilization for a wide range of inference applications. 

\begin{figure}[t]
  \centering
  \includegraphics[width=0.8\linewidth]{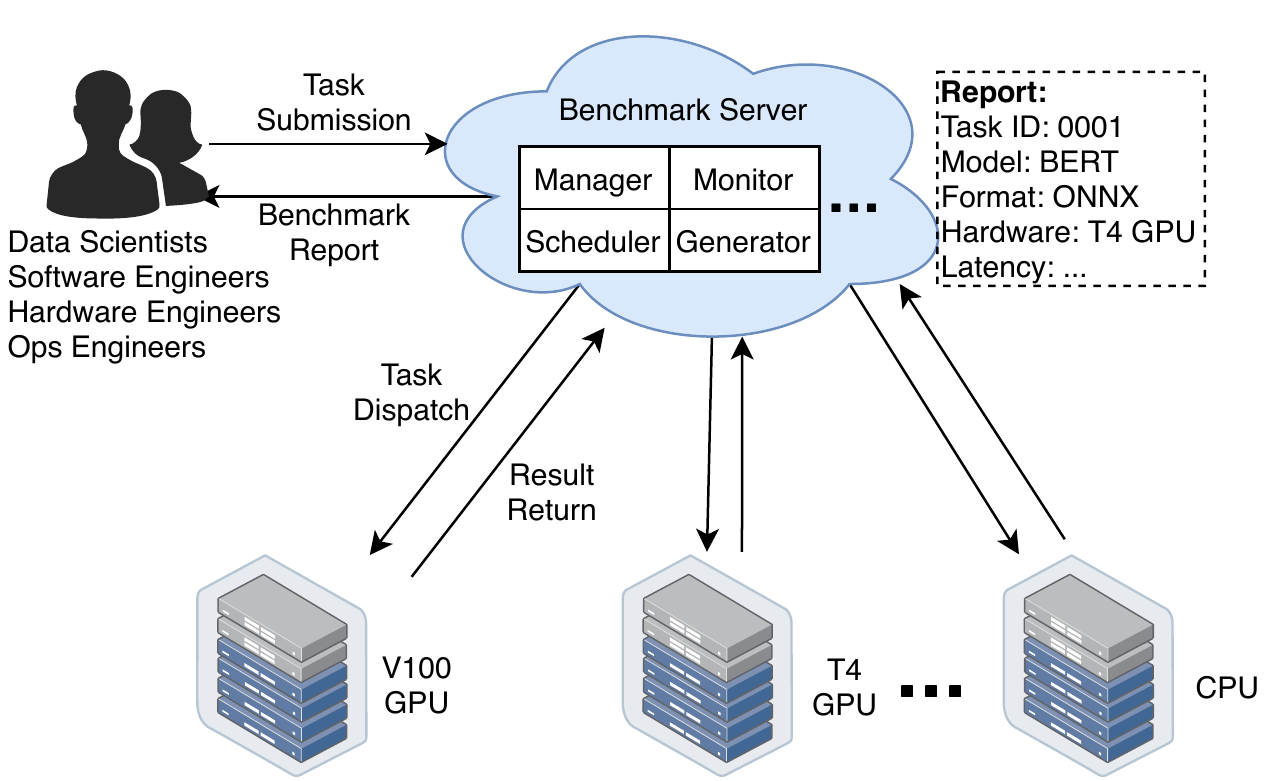}
  \caption{The overview of the proposed benchmarking system. The system first accepts users' benchmarking tasks. Then it distributes the tasks to dedicated servers to complete them automatically. Finally, it will send a detailed report and guidelines back to users.}
  \label{fig:benchmark_server}
\end{figure}

In this work, we propose an automatic and complete DL serving benchmark system (as shown in Figure \ref{fig:benchmark_server}) to address these needs and narrow gaps of existing systems. We design our system following the best practice in DL clusters design. The goal is to provide an end-to-end solution that frees DL developers from tedious and potentially error-prone benchmarking tasks (e.g., boilerplate code writing, data collection and workload generation). In addition to the models chosen by users, the system can effortlessly generate and iterate models with different hyper-parameters (e.g., different layer types and different number of layers) to adequately explore the design space. The system also provides an extensive set of analysis tools and models to help developers choose the best configuration for their applications under constraints like latency, cloud cost, etc. Besides, we implement a two-tier scheduler in the system to improve service efficiency.


For the logical clarity and comprehensiveness of benchmarking evaluations, we designate impact factors for DL inference into three tiers, as shown in Figure \ref{fig:benchmark_tier}. For the hardware tier, we select five representative platforms for evaluation. For the software tier, we choose four representative online serving infrastructures. For the pipeline tier, we simulate the real-world workload and examine a specific inference service with three types of transmission technologies. We use metrics such as tail latency, cloud cost and resource usage to measure their performance, and analysis models like Roofline \cite{williams2009roofline} and Heat maps to profile DL serving system properties.

In summary, the main contributions of this paper are as follows,
\begin{itemize}
    \item We build an automatic and distributed benchmark system in DL clusters to streamline DL serving inference benchmark for developers.
    \item We conduct a comprehensive performance analysis on three tires with many DL applications under various workloads, providing users insights to trade off latency, cost, and throughput as well as configure services.
    \item We use generated models to study the sensitivity of hardware performance to model hyper-parameters and derive valuable insights for system design.
    \item We implement a scheduler to ensure safe benchmark progress and reduce the average benchmark job completion time.
\end{itemize}

In the remainder of this paper, we first introduce the background knowledge of DL inference serving and our motivation in Section \ref{sec:background}. Then, we detail three benchmark tiers as well as their benchmarking metrics in Section \ref{sec:benchmark_tier}. Next, we present the system implementation and the employed methodologies in Section \ref{sec:system_design}. We employ our system to perform benchmark tasks and evaluate its performance in Section \ref{sec:result}. Finally, we briefly introduce the related work in Section \ref{sec:relatedwork} and summarize our paper in Section \ref{sec:summary}.

\section{Background and Motivation}
\label{sec:background}
In this section, we introduce the typical workflow of building DL services including pre-deployment, post-deployment and online serving. We illustrate how current benchmark studies can not fully address the challenges of the new workloads, thus motivating us to build a new benchmark system.

\subsection{Deep Learning Service Pre-Development}

\begin{figure}[t]
  \centering
  \includegraphics[width=0.95\linewidth]{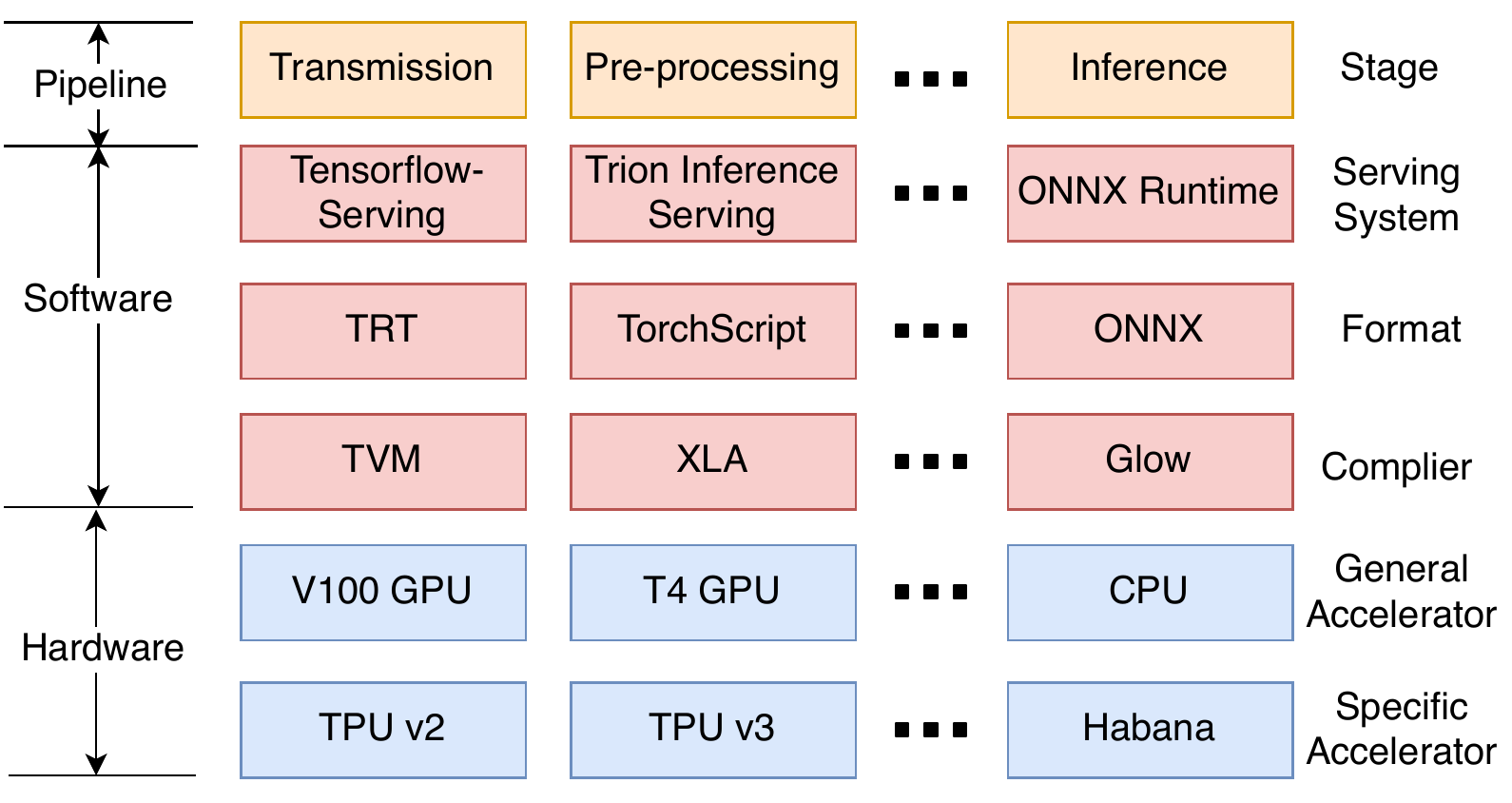}
  \caption{Three benchmark tiers for evaluating DL inference. The performance (e.g., latency and cost) of an AI application is influenced by many factors, which can be categorized into three classes: hardware, software, and pipeline.}
  \label{fig:benchmark_tier}
\end{figure}

\begin{figure}[b]
  \centering
  \includegraphics[width=0.95\linewidth]{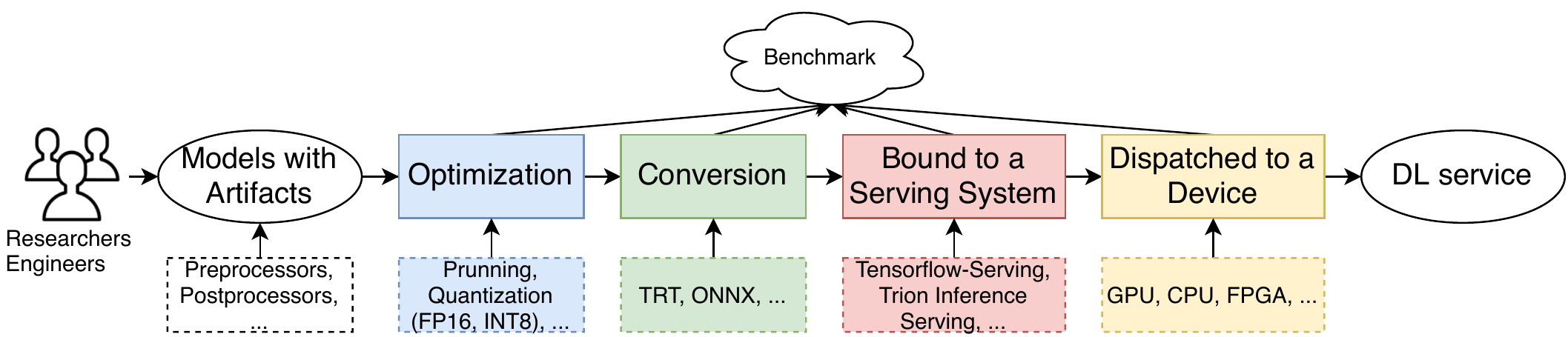}
  \caption{The pipeline of the AI service deployment. To build a high-performance and low-cost AI service, users need to spend a lot of time on optimization and benchmarking.}
  \label{fig:ml_deployment_workflow}
\end{figure}

After receiving models from data scientists, engineers need to perform many optimization and service configuration tasks before models go to production, as shown in Figure \ref{fig:ml_deployment_workflow}. First, models alone with their artifacts such as weight files and corresponding processors (e.g., image resize), will be stored and versioned \cite{chard2019dlhub}. Developers may try to re-implement processors with a production language such as Java and \textbf{test} them. Second, engineers will optimize models (e.g., INT8 conversion or model compression \cite{han2015deep}) but maintaining good accuracy. As such, engineers need to \textbf{check} every new model to ensure that it meets both accuracy and latency requirements. Third, the users will choose a serving system such as Tensorflow-Serving (TFS) \cite{baylor2017tfx} to bind their models as a service. Different serving systems perform differently over a wide range of hardware choices. Even more, theses systems have many new features such as dynamic batching \cite{crankshaw2017clipper}, resulting in a complex configuration with varied performance. Developers need to perform rigorous \textbf{evaluations} for a service. Finally, a service can be dispatched to a cluster to serve customers. 

\textbf{Observation 1:} A benchmark task for building a service often takes tens of iterations. Also, developers need to consider the trade-off among many impact factors (IFs) under SLOs (Service-Level-Objectives) or budget to make a judicious choice. To efficiently complete the task, developers need a simple, easy-to-use and customizable system.

\textbf{Observation 2:} The benchmark tasks need a dedicated and isolated runtime environment. As AutoML techniques are becoming popular, these tasks will consume more resources, resulting in long delays in getting benchmark results. This motivates us to propose a way to perform a benchmark with efficient use of resources.

\subsection{Deep Learning Service Post-Development}

Upon finishing the pre-development, the developers still need to 1) study the accelerator (e.g., TPU \cite{jouppi2017datacenter} and FPGA) characteristic for upgrading them; 2) design better resource allocation mechanisms for DL serving in clusters; 3) monitor DL services to diagnose performance issues immediately. The automation of these tasks has been studied extensively. For instance, while many engineers manually write DL operators, TVM \cite{chen2018tvm} automatically generate operators to run models on specific hardware efficiently. Also, to increase the GPU resource utilization, MPS \cite{nvidiamps} and Salus \cite{yu2019salus} provide supports to share a GPU among multiple models. All of these studies require a lot of benchmark efforts. However, current benchmark studies ignore these post-deployment activities and provide very few supports for developers.

\textbf{Observation 3: } Resource usage under different scenarios is not fully studied, which limits the development of resource allocation methods for DL services. Developers need to analyze resource usage with varied settings to improve resource utilization.

\textbf{Observation 4: } Current studies for simple isolated models can not be easily generalized to analyze system bottleneck (e.g., compute- or memory- bound). Also, only benchmarking these models can not provide help to understand the hyper-parameter influence on both DL hardware and model performance. More effective analysis methodologies should be incorporated.

\subsection{DL Inference Serving Workflow}

\begin{figure}[t]
  \centering
  \includegraphics[width=0.95\linewidth]{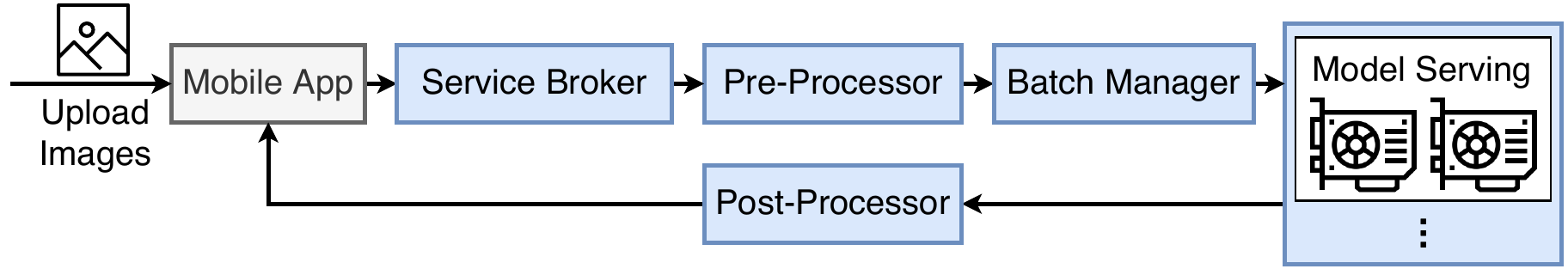}
  \caption{A simplified view of a DL inference pipeline. Users first send requests through a mobile or web app. A service broker then distributes requests to backend servers for processing. Usually, during the processing, a request will go through a pre-processor, a batch manager, a model server and a post-processor.}

  \label{fig:probe_pipeline}
\end{figure}

In this section, we depict how a DL Service process users' requests and point out the limitations of current studies. As shown in Figure \ref{fig:probe_pipeline}, customers first send requests from mobile or website. These requests will be pre-processed either on the client- or server-side to meet a format requirement of a model service. The requests are then forwarded to a frontend in a server for further dispatched. Next, the request will be fed into a backend where a model or multiple models is loaded by a serving platform (e.g., TFS). Once the inference is completed, the predictions will be post-processed (either in a server or a client) and displayed to the end-users.

\textbf{Observation 5:} The end-to-end design of the pipeline determines the performance of a service. As such, it is very useful to have a detailed benchmark to locate the bottleneck.

\textbf{Observation 6:} Since a service processes hundreds of thousands of requests daily \cite{hazelwood2018applied}, the mitigation of tail latency under varied arrival rates, is critical. Serving software usually includes functions to address the issue. Developers should have a tool to analyze this.

\textbf{Observation 7:} Existing DL serving systems \cite{crankshaw2017clipper, baylor2017tfx} provides several mechanisms such as dynamic batching to improve the performance of a service. A benchmark system should explore these advanced features.

\section{Benchmark Tiers}
\label{sec:benchmark_tier}

In this section, we describe three benchmark tiers: hardware, software, and pipeline as shown in Figure \ref{fig:benchmark_tier}.

\subsection{Tier 1 - Hardware}
\label{sec:tier_hardware}

For the first tier, our system mainly focuses on inference hardware performance as well as the interactions between hardware and models under varied configurations and scenarios. In addition to the latency and throughput measurement capabilities provided in previous research, we extensively study the cost, hardware performance sensitivity to model hyper-parameters, and their bottleneck with different types of models. We next detail the metrics.

\textbf{Latency \& Throughput. } These two metrics are widely used to measure hardware. In general, online real-time services require low latency and offline processing systems prefer high throughput. Unlike CPUs, the new accelerators such as TPUs and GPUs encourages batch processing to improve resource utilization. We extensively studied the property with our system.

\textbf{Cost.} When developers implement a DL service, they must consider their budget. To support this, our system provides tools to measure energy, CO2 emission \cite{anthony2020carbontracker}, and cloud cost under different devices and cloud providers with an aim to present a comprehensive study.

\textbf{Sensitivity of Model Hyper-parameters.} DL models have many hyper-parameters such as layer types (e.g., LSTM \cite{hochreiter1997long}) and the number of layers. Our system can generate models to study their influences to provide insights.

\textbf{Memory \& Computation. } The performance of a model on a device is decided by both computation and memory capacity from the device. We explore both of them with real-world and generated models under a wide range of hyper-parameters.


\subsection{Tier 2 - Software}

In the second tier, we study the impacts of software such as formats and serving platform. Here, the system provides complete support for serving platforms such as TFS \cite{olston2017tensorflow} and Trion Inference Serving \cite{trtserving}, which have not been fully investigated in the previous study. Users can easily extend our systems to support more platforms with provided APIs. We will study the following features.

\textbf{Tail Latency.} A well-designed serving platform can effectively mitigate the effect of tail latency which is critical for online services. We provide a detailed study of the performance under varied request arrival rates.

\textbf{Resource Usage. } Serving software may bring overhead or save resources by their dedicated design. Understanding the behavior will lead to better resource allocation.

\textbf{Advanced Features. } To improve resource utilization, serving software provide some useful functions like dynamic batching. Our system can help to study their impacts.

\subsection{Tier 3 - Pipeline}

In this tier, we aim to look into each stage to identify the bottleneck of a service under various conditions. The system supports to explore the performance of each stage.

\textbf{Latency per Stage.} The DL serving pipeline often consists of five stages: pre-processing, transmission, batching, DL inference, and post-processing. In addition to the DL inference stage, the other stages also have varied performance under different conditions (e.g., transmission under varied networking conditions). We will discuss this in detail. Meanwhile, the cold start that refers to the time interval (milliseconds to seconds) to start a system is varied for different systems. It is also a critical metric as it decides the provisioning time.

\textbf{Summary.} The huge configuration space leads to many trade-offs including but not limited to \textbf{Latency versus Throughput, Accuracy versus Latency, Cost versus Quality and Sharing versus Dedicate}. With the help our system, DL developers can inspect to these trade-offs for better configuration and future upgrading.

\section{System Design and Implementation}
\label{sec:system_design}



\textbf{Design Philosophy.} We aim to design an automatic and distributed benchmark system which can be 1) operated independently to perform benchmark tasks; 2) incorporated into DL lifecycle management \cite{zaharia2018accelerating} or DL continues integration \cite{zhang2020mlmodelci, karlavs2020building} systems to further improve the automation of DL service development and evaluation; 3) and connected to a monitor system for DL service diagnose. Accordingly, our system can serve users involved at various stages of a DL service pipeline: data scientists, DL deployment engineers, hardware engineers, and AIOps engineers, by significantly reducing the often manual and error-prone tasks.

\subsection{System Overview}
\label{system_overview}

\begin{table*}[t]
\centering
\caption{Five hardware platforms we used for experiments.}
\label{tab:hardware_spec}
\begin{adjustbox}{width=1\textwidth}
\begin{tabular}{cccccccc}
\hline
\textbf{ID} & \textbf{\begin{tabular}[c]{@{}c@{}}Platform \\ (Arch)\end{tabular}} & \textbf{Version}      & \textbf{Memory} & \textbf{\begin{tabular}[c]{@{}c@{}}Peak TFLOPS\\ (FP32/FP16)\end{tabular}} & \textbf{\begin{tabular}[c]{@{}c@{}}Memory\\ Bandwidth (GB/s)\end{tabular}} & \textbf{\begin{tabular}[c]{@{}c@{}}AWS\\ (Instances)\end{tabular}} & \textbf{\begin{tabular}[c]{@{}c@{}}Google Cloud\\ (Instances)\end{tabular}} \\ \hline
C1          & CPU                                                                 & Intel Xeon E502698 v4 & 128 GB              & -                                                                          & -                                                                          & -                                                                  & -                                                                           \\ \hline
G1          & GPU (Volta)                                                         & Tesla V100            & 32 GB           & 15.7 (31.4)                                                                & 900                                                                        & 4                                                                  & 4                                                                           \\ \hline
G2          & GPU (Turing)                                                        & GeForce 2080Ti        & 11 GB           & 14.25 (28.5)                                                               & 616                                                                        & -                                                                  & -                                                                           \\ \hline
G3          & GPU (Turing)                                                        & Tesla T4              & 16 GB           & 8.1 (16.2)                                                                 & 300                                                                        & 7                                                                  & 3                                                                           \\ \hline
G4          & GPU (Pascal)                                                        & Tesla P4              & 8 GB            & 5.5 (11.0)                                                                 & 192                                                                        & -                                                                  & 3                                                                           \\ \hline
\end{tabular}
\end{adjustbox}
\end{table*}

As shown in Figure \ref{fig:benchmark_server}, our centralized benchmark system uses a Leader/Follower architecture. The leader server manages the whole system by accepting users' benchmark submissions and dispatching benchmark tasks to specific follower workers, guided by a task scheduler. The leader server also generates the corresponding requests and workloads for later evaluation. Meanwhile, users can choose either to register their own models to our system or use the different iterations of canonical models generated by our system to benchmark. Next, the follower workers, which can be any of the idle servers in a cluster, execute users' benchmark tasks according to the submissions' specifications. Specifically, it loads and serves a model with infrastructure such as TensorFlow-Serving (TFS) and then invokes the corresponding pre-processing and post-processing functions to complete a service pipeline. Finally, users can start clients to simulate real-world workload by sending requests to the DL service to evaluate its performance.

\begin{figure}[t]
  \centering
  \includegraphics[width=\linewidth]{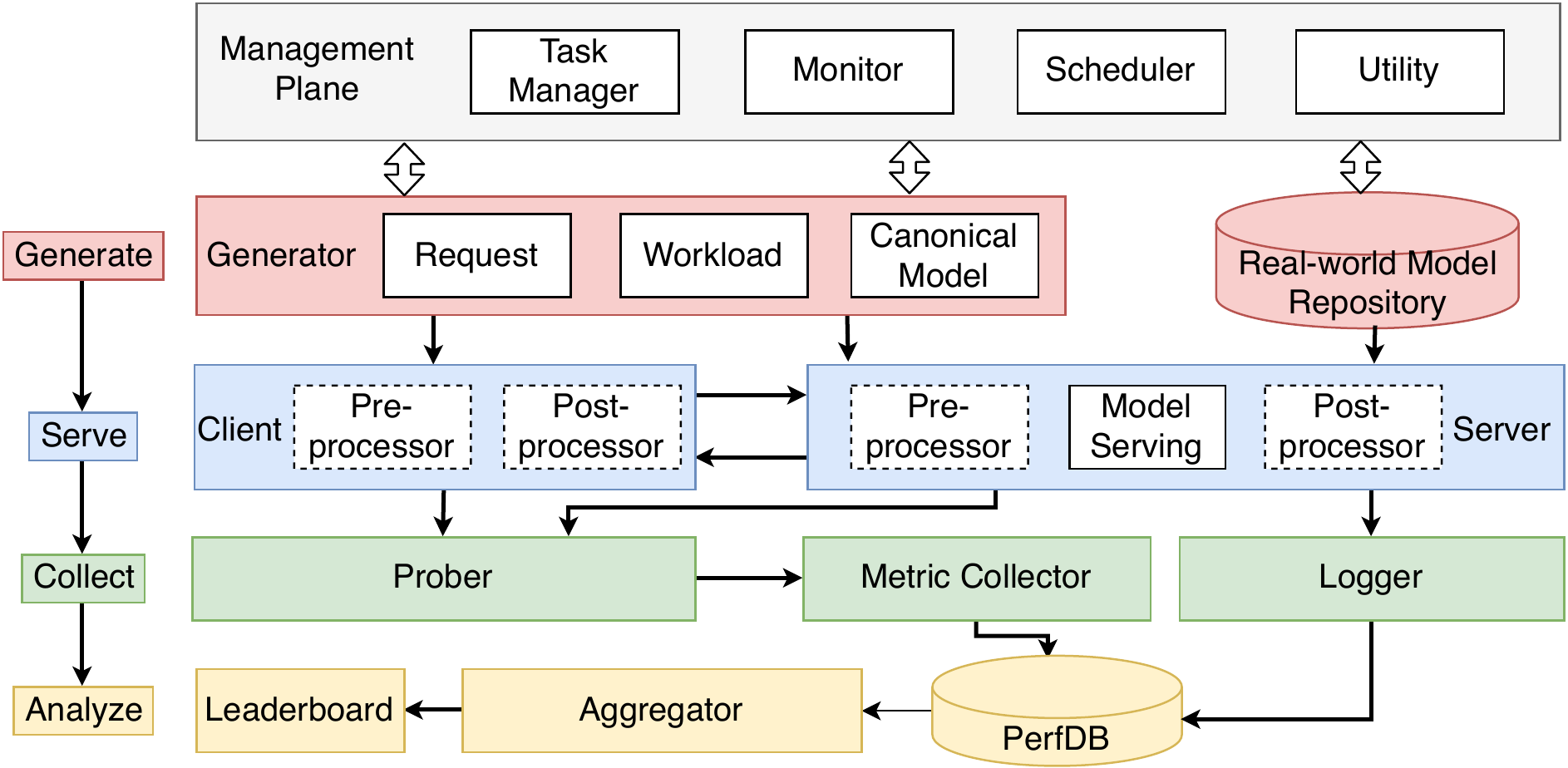}
  \caption{The benchmarking system architecture. A management plane running in the leader server is designed to control and monitor the system and users' benchmarking tasks. We divide an inference benchmarking into four stages and provide a set of functions to complete it automatically.}
  \label{fig:system_overview}
\end{figure}

For evaluation, a metric collector is utilized as a daemon process to obtain detailed performance information via a probe module. At the same time, a logging module is enabled to record the running systems' status, including both hardware and software. All of this information will be sent to a PerfDB (performance database). After evaluation, users can aggregate the performance data and use our build-in analysis models to extract insights. We also implement a recommender and a leaderboard for service configuration and resource allocation.

\subsection{System Implementation Detail}
\label{implementation_detail}

Our implementation of the system (shown in Figure \ref{fig:system_overview}) follow the following three best-practices:

\textbf{Modularity.} Modularity can provide 1) a seamless extension to support evolving DL applications, 2) a natural adaptation to existing model management and serving systems, and 3) easy customization from developers.

\textbf{Scalability.} As cluster-based model training and deployment have become the de facto practice for both industry and academia, our system supports all the computation and management functionalities required by cluster computing, with the aim to keep user manual efforts (e.g., manually moving model and result files) to the minimum.

\textbf{System Integrity and Security.} Since starting a benchmarking task without proper coordination with the leader can potentially interrupt the running service, the system shall have the capability to monitor environment status so as to decide whether a task can be executed or to schedule such a task.

\subsubsection{Management Plane}

The management plane, designed to control all benchmark tasks in users' clusters, consists of four functional blocks: task manager, monitor, scheduler, and other utility functions. The goal is to provide an interface for users' fine-grained control inputs.

\textbf{Task Manager.} The task manager accepts users' benchmark submission and logs related information, such as the user name, task ID, and submission timestamp. The task manager then dispatches the task to a specific worker according to the task specification and starts the related procedures. In our implementation, we use MongoDB as the backend. The developers can easily replace the backend with their preferred database such as MySQL.

\textbf{Monitor.} This functional block collects and aggregates system resource usage. Specifically, We use two backends, cAdvisor \cite{cadvisor} and Node Exporter \cite{nodeexporter}, for logging the status of serving container (e.g., CPU usage) and hardware status (e.g., GPU usage), respectively.

\textbf{Scheduler.} We aim to design a multi-tenant system to organize a team's daily benchmark tasks and avoid potential conflict and interference. As such, we implement a scheduler for these benchmark jobs to minimize the average job completion time (JCT) and improve efficiency. Specifically, we design a simple baseline method (as discussed in Section \ref{sec:scheduler}), upon which developers can extend to design their own algorithms per the workload profiles.

\textbf{Utility Functions.} Currently, we have two functions, a sharing manager and a configuration recommender. The sharing manager helps users configure MPS, which is the de facto software to share multiple DL models with one GPU, to support a sharing benchmark. The recommender help users to make a simple decision for service configuration. Users need to input an SLO (e.g., latency), and the system will return the top 3 configurations.

\subsubsection{Stage 1 - Generate}

In the first stage, our system prepares requests, workloads (sending patterns), and models for benchmarking according to users' specifications. From their submission (a \textit{YAML} file), the system first chooses to call either real-world or generated models and then prepares the corresponding requests and workloads to perform tasks. We next describe the four main functions of this stage.

\textbf{Model Repository.} As each team can produce tens or even hundreds of models, this module is designed to help them organize different versions of models. It has four APIs, including \textit{register}, \textit{update}, \textit{search}, and \textit{delete}, based on MongoDB with GridFS. Both model weights and basic information such as the model name and the dataset used can be stored in the repository.

\textbf{Canonical Model Generator.} The module contains a trainer and an exporter to generate canonical models under different hyper-parameter, including the batch size, the neuron number, and the layer number (see appendix for details). We build models by repeatedly stacking the following four most commonly used blocks (layers), a fully-connected layer (FC), a residual block (CNN), an LSTM layer (RNN), and an attention block (Transformer \cite{vaswani2017attention}), respectively, to obtain four groups of models. They are trained with Tensorflow and exported to a format that can be served by Tensorflow-Serving. Different from those isolated real-world models registered by users, the canonical models can help explore the sensitivity of hardware performance to model hyper-parameters, which can not only benchmark target platforms but also profile their properties.

\textbf{Request Generator.} To save developers' preparation time, the module stores many kinds of data selected from widely used datasets such as ImageNet \cite{deng2009imagenet} and also has an interface for users to upload their own test data.

\textbf{Workload Generator.} Since the requests must be sent by following a pattern for benchmarking, we implement this workload generator. We have built many modes in our system to meet the diversity testing scenarios (including both online services and offline services). Developers can further customize all these modes. For instance, we have a pattern to simulate request arrival processes that follow a Poisson Distribution and a specified arrival rate.

\subsubsection{Stage 2 - Serve}
\label{sec:serve}


In the second stage, the system performs users' benchmark tasks according to the schedule. To expedite the execution, we implement several unified functionalities, including a pre-processor, a post-processor, and a model server.

\textbf{Pre-processor \& Post-processor.} We collect and implement many out-of-the-box processing functions for different DL models in our system. For instance, for image classification models, we have image resizing and tensor conversion functions. For text classification models, we provide a set of tokenizer methods. These functions can be called individually or jointly according to model specifications. For instance, when analyzing videos, we need to build a pre-processing pipeline that consists of video decoding and image resize functions. The post-processor also has a variety of functionalities. For example, the post-processor can match a prediction class ID to a label in a database for classification tasks.

Following our design, developers can extend the system by implementing their own processors. Our system also provides the flexibility for users to offload these functions to the client-side or directly run them on the server-side.

\textbf{Model Server.} The benchmarking of DL serving software infrastructure has not been thoroughly investigated by the current benchmark studies. As such, we provide the module to bridge the gap. The serving software can transform models into services and expedite the model development. We adopt two kinds of DL serving infrastructure into our system. The first is the DL-specific software, with Tensorflow-Serving (TFS) and Trion Inference Serving (TrIS) as two representatives. The second is a general web framework (e.g., FastAPI and Flask) with an optimized runtime (e.g., ONNX Runtime \cite{onnxruntime}). We wrap many of them in our system so that developers can start a service for benchmarking with ease.

\subsubsection{Stage 3 - Collect}

In this stage, three modules are implemented to ensure both fine-grained performance collection and reproducibility. All of the information will be sent to the performance database.

\textbf{Prober.} The module examines the DL serving pipeline to obtain the performance of each stage. As shown in Figure \ref{fig:probe_pipeline}, to evaluate the performance of a serving pipeline with multiple stages, the prober sets endpoints at the boundaries of each stage. Then it triggers the metric collector to obtain the measurement results corresponding to each of them. The results help developers detect the performance bottlenecks in an inference pipeline.

\textbf{Metric Collector.} This module contains a series of evaluation metrics as well as their implementation, as described in Section \ref{sec:benchmark_tier}. This module is designed to be composable - it can either be fully automated during the benchmarking process or be called selectively to meet users' demand.

\textbf{Logger.} This module records the system runtime information and parameter settings. As we discussed in Section \ref{sec:benchmark_tier}, the DL inference performance is affected by model, hardware, input size, etc. To ensure the benchmarking results' reproducibility, we use a logger to track all the above information during evaluations. The information falls into two categories: runtime environment information and evaluation settings. Runtime environment information includes hardware types, serving software names, etc. Evaluation settings include model names, training frameworks, etc.

\subsubsection{Stage 4 - Analyze}

In the last stage, users can get initial benchmark results from a database and use analysis models and tools built in our system to gain insights. A leaderboard is provided for users to check results.

\textbf{PerfDB.} We design a performance database to store benchmark results and log information from the collection stage. The database runs a daemon process in the leader server to collect data from those follower workers in a cluster. Same with the model repository, it exposes many functions for results management MongoDB as the backend.

\textbf{Aggregator.} The module is used to aggregate results from PerfDB for a comprehensive analysis. We adopt many analysis models (e.g., Roofline) to our system to draw more insightful conclusions. The models will be described in Section \ref{analysis_model}.

\textbf{Leaderboard.} In the leaderboard, developers can sort results by different metrics such as energy computation and cloud cost to configure a high-performance and cost-effective DL service. We also provide a visualizer for presenting the results.

\subsection{System Methodology}
\label{analysis_methodology}

This section introduces adopted analysis models in our system and a simple scheduling method for benchmark tasks.

\subsubsection{Analysis Model}
\label{analysis_model}

\textbf{Roofline.} Roofline \cite{williams2009roofline} model is widely used to study the performance of applications running on hardware devices. We use Roofline to estimate two core capacity metrics of an accelerator: computation and memory bandwidth. Compared to a simple percent-of-peak estimate, it evaluates the quality of attained performance and explores a performance bound more effectively. As DL models' performance is affected by many hyper-parameters (e.g., neuron numbers), we use generated canonical models to exploit the performance as well as the interactions between hardware and models instead of simple real-world models.

\textbf{CDF Plots.} Cumulative distribution function (CDF) plots can give the probability that a performance metric (e.g., latency) is less than or equal to its target (e.g., latency SLO). With the plot, we compare the capability to process real-world workloads of varied software.

\textbf{Heat Maps.} Heat maps are powerful tools to understand the performance (e.g., utilization) sensitivity to models' hyper-parameters. We rely on heat maps to measure how system performance varies with the hyper-parameters to help developers better understand the interactions between models and systems.

\textbf{Other Plots.} We also design and upgrade some basic bar plots to summarize the performance obtained from our system. Compared to a table that lists results, our plots highlight the insights and give intuitive guidelines for service configuration.

\subsubsection{Scheduler Design}
\label{sec:scheduler}

Our system invokes a scheduling agent in a two-tier manner to perform benchmarking tasks efficiently. In the first tier, a newly submitted job will be dispatched by the leader server to a relatively idle follower worker with adequate resources to execute the job. The second tier is to determine the order of job execution on a follower worker. Specifically, suppose the system has a set of jobs $J$ waited to be processed in a scheduling interval. For each job $j \in J$, the total time to process a job is $t_j = waiting + processing$. The optimization target is to minimize $T$ where $T = \sum_{j \in J} t_j$.

\begin{algorithm}[h]
   \caption{Benchmark Job Scheduling}
   \label{alg:scheduler}
\begin{algorithmic}
   \STATE {\bfseries Input:} Jobs $J$, Workers, $W_1, \dots, W_k$
   \FOR{all jobs $j \in J$}
   \REPEAT
   \STATE Select an idle worker $W_{min}$ with the shortest queue
   \STATE Enqueue $j$ to $W_{min}$
   \STATE Remove $j$ from $J$
   \UNTIL{$J$ is empty}
   \ENDFOR
   \FOR{all Workers, $W_1, \dots, W_k$}
   \STATE Re-order jobs in an ascending way
   \ENDFOR
   \STATE Execute Jobs
\end{algorithmic}
\end{algorithm}

To solve the problem, we implement a global load balancer (LB) to decide the job placement and a scheduler at every follower worker to allocate resources, as shown in Algorithm \ref{alg:scheduler}. First, workers will publish their current queue length (i.e., the time to process all waited jobs) to the leader server. Then LB distributes a job to a worker, minimizing the waiting time. Next, the worker will re-order jobs in an ascending way. Finally, workers will execute the jobs sequentially.

\section{Evaluation Results}
\label{sec:result}

In this section, we describe the experimental settings and the evaluations of our system. We also illustrate the importance of benchmarking scheduling via a simple case study.

\subsection{Experimental Setup}
\label{exp_setup}

We deploy our system to a server in a private cluster and study five types of hardware, four types of software platforms, and a pipeline with three scenarios. In the process, we have registered many real-world models and generated hundreds of models for the benchmark study.

\textbf{Hardware Platforms.} The specifications of five hardware platforms are listed in Table \ref{tab:hardware_spec}, including one CPU platform as the reference and four GPU platforms (V100, 2080Ti, T4, and P4). These GPUs have many different architecture designs (e.g., Volta and Turing), thus providing us a wide range of computational capabilities. We also survey instances that provide the GPUs in cloud providers.

\begin{figure}[t]
  \centering
  \includegraphics[width=0.8\linewidth]{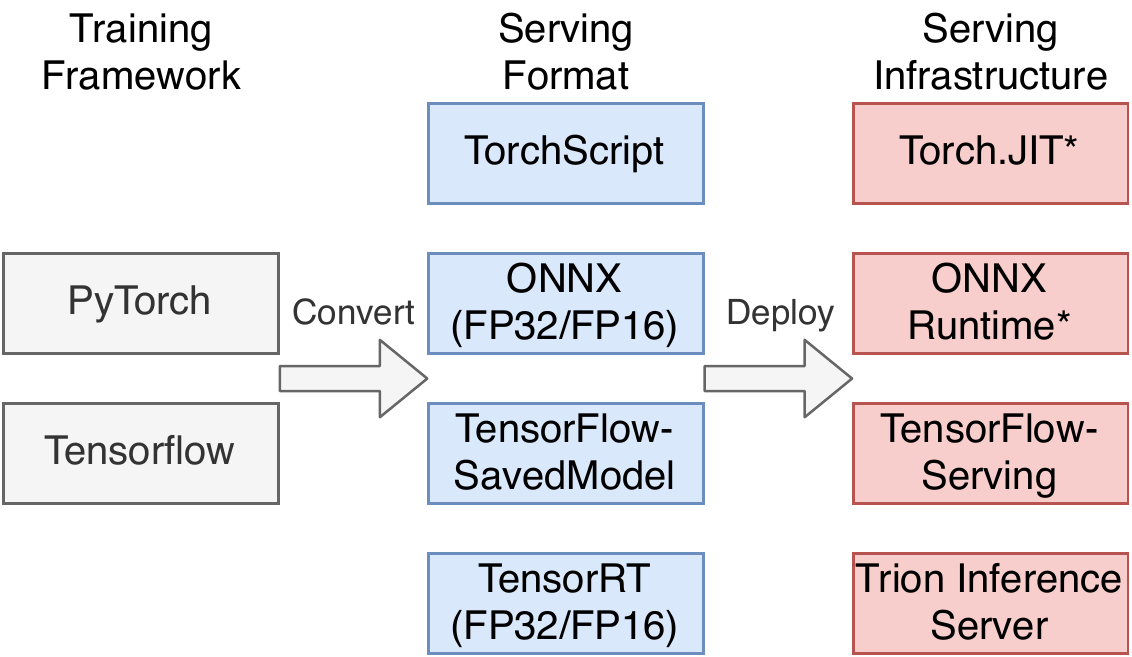}
  \caption{Four serving software infrastructures under our test. Our system accepts models trained with PyTorch and Tensorflow. Then it converts the models into many serialized and optimized serving formats such as ONNX and TensorRT. These serving model will be bound with a serving infrastructure like Tensorflow-Serving for benchmarking.}
  \label{fig:serving_software_infra}
\end{figure}

\textbf{Software Platforms.} As shown in Figure \ref{fig:serving_software_infra}, we investigate four serving infrastructures in our experiments. Tensorflow-Serving (TFS) is the default serving platform for Tensorflow SavedModel format, which is converted from Tensorflow models. Trion Inference Server (TrIS) supports many formats like TensorRT format that can be converted from many other formats. Both TorchScript (converted from PyTorch) and ONNX (converted from PyTorch) have no stable serving infrastructure during our implementation, so we use their default optimized runtime torch.jit and ONNX runtime, respectively, to load models and wrap them as services with FastAPI (a popular web framework \cite{fastapi}) by following the official document. To use these platforms, our system first converts newly trained models into optimized and serialized formats. We next use gRPC APIs test them with docker containers. All of them have been adopted to our system to free developers from engineering works.

\textbf{Pipelines.} We build a pipeline (shown in Figure \ref{fig:probe_pipeline}) that consists of a service broker that accepts users' requests and dispatches them to a backend with a running DL service, as our testbed in the system. We test the whole pipeline in three network scenarios, including LAN, 4G LTE, and Campus WIFI.

\subsection{Hardware Platform Characterization}
\label{hardware}

We present hardware benchmarking results with both real-world and generated models in this section. More results can be found in the leaderboard at our website (we omit it here due to the anonymous review). Unless otherwise stated, all experiments are conducted with TensorFlow SavedModel with TFS 2.3.0.

\begin{figure}[ht]
\begin{subfigure}{0.5\columnwidth}
  \centering
  \includegraphics[width=1.0\linewidth]{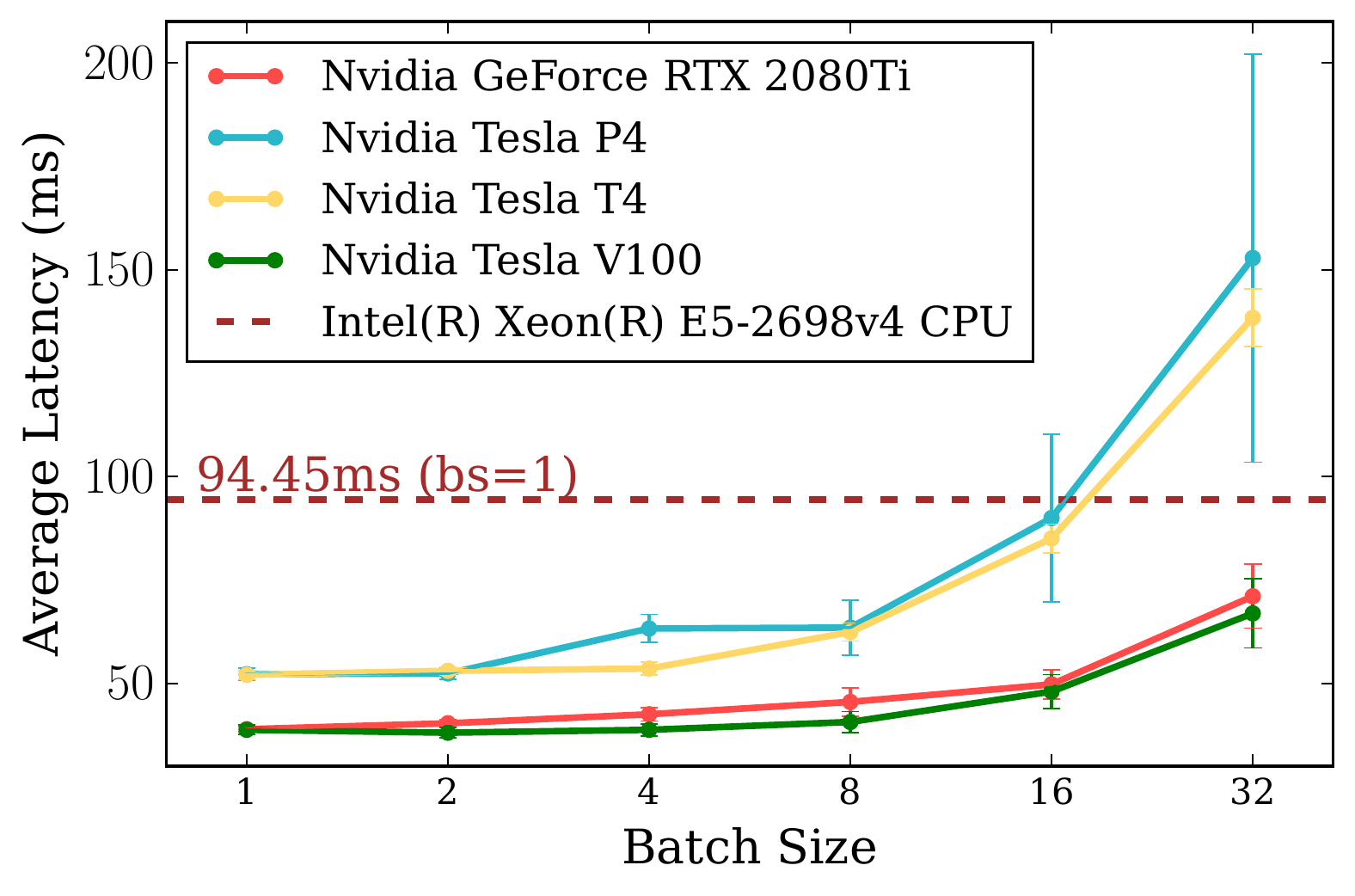}
  \caption{Bert-Large}
  \label{fig:hardware_bert_latency}
\end{subfigure}%
\begin{subfigure}{.5\columnwidth}
  \centering
  \includegraphics[width=1.0\linewidth]{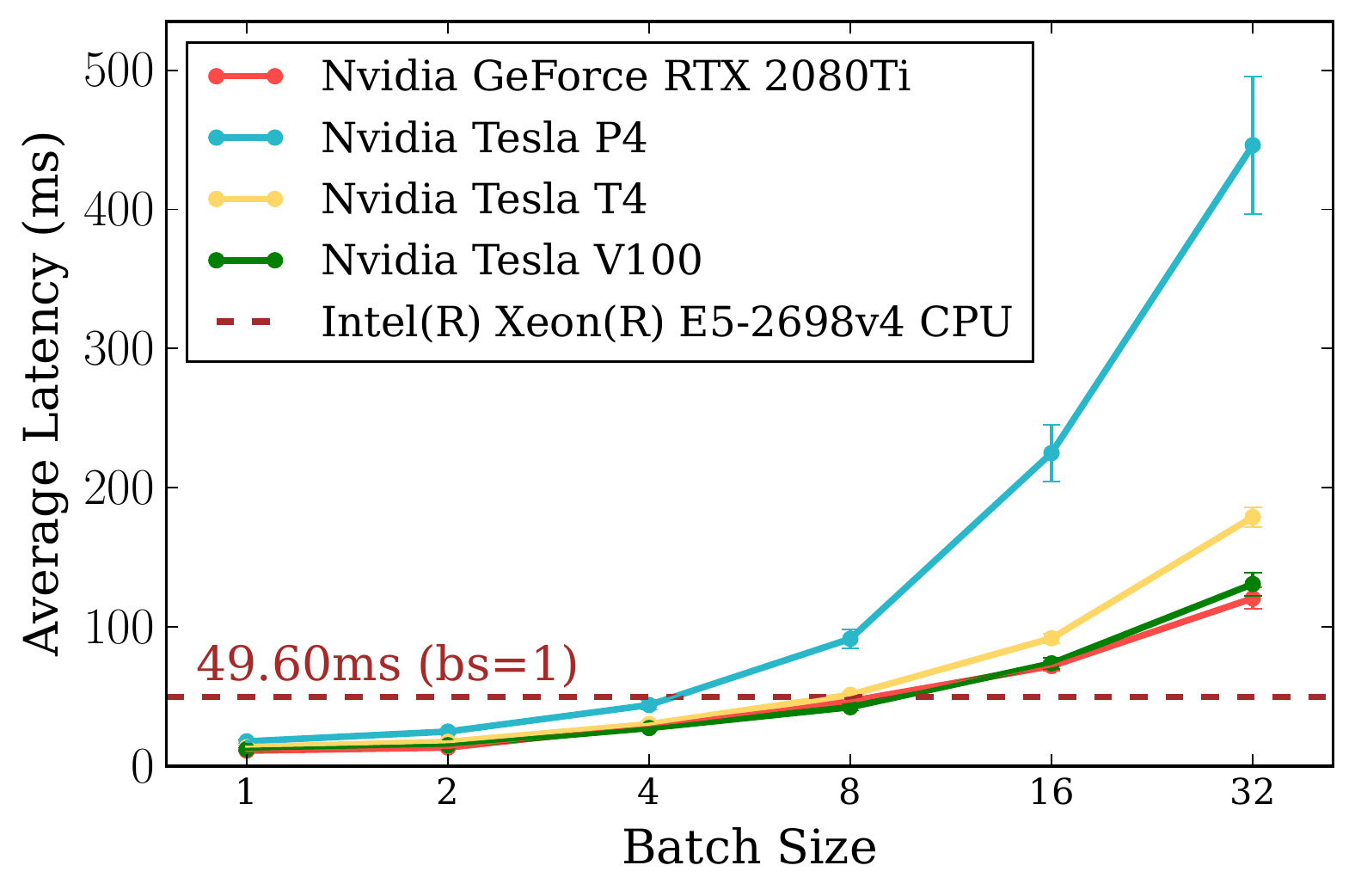}
  \caption{ResNet50}
  \label{fig:hardware_resnet50_latency}
\end{subfigure}
\begin{subfigure}{1.0\columnwidth}
  \centering
  \includegraphics[width=1.0\linewidth]{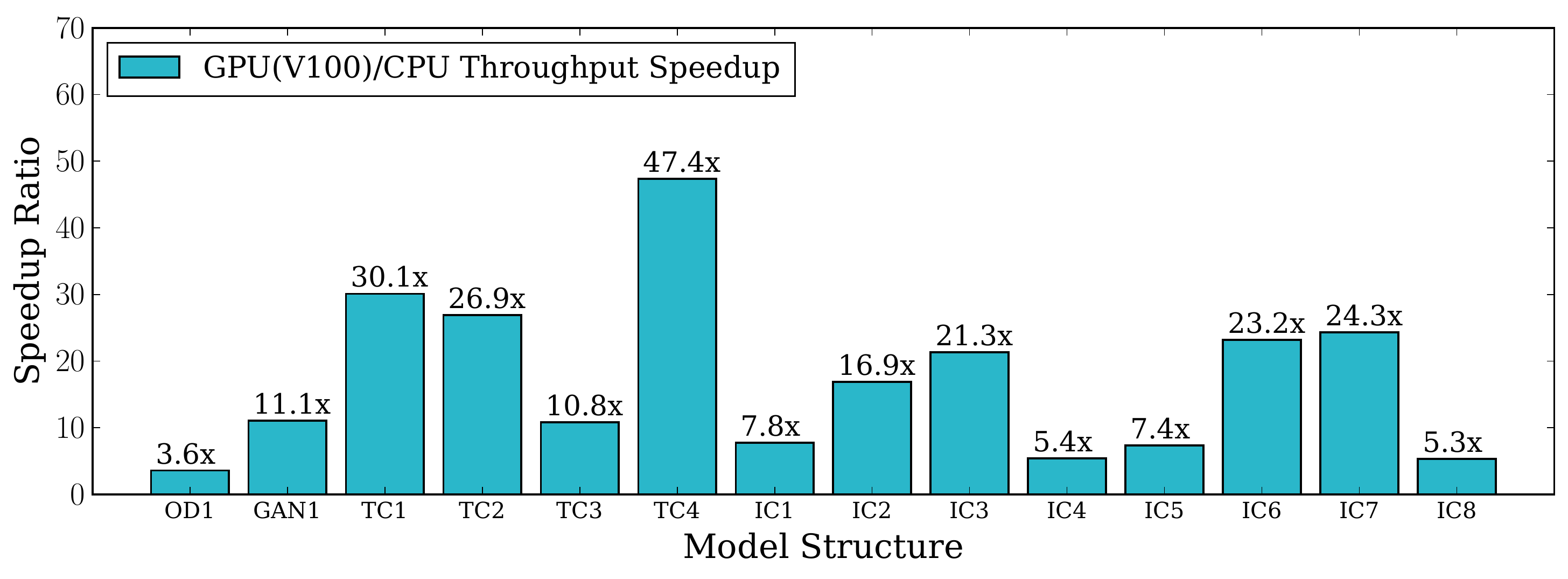}
  \caption{GPU/CPU speedup for ResNet50}
  \label{fig:hardware_throughput_speedup}
\end{subfigure}
\caption{The latency and throughput comparisons of different hardware. We showcase two representative models with varied batch sizes. We also show the speedup of GPU/CPU under the latency SLO.}

\label{fig:hardware_latency_throughput_speedup}
\end{figure}

\textbf{Latency \& Throughput.} We first plot how latency changes with different batch sizes and type of hardware for two inference models (Bert-Large and ResNet50), in Figure \ref{fig:hardware_bert_latency} and \ref{fig:hardware_resnet50_latency}, respectively. The batch size for the CPU is fixed at one. The plots show that for latency, GPU platforms perform better than CPU for small batch sizes (less than eight). When the batch size becomes large, the latency becomes much longer for two types of GPU. A larger batch size often provides high throughput. We caution developers to check they latency SLO first and use our configuration recommendation to select hardware and batch size before deploying the service. In Figure \ref{fig:hardware_throughput_speedup}, we plot the speedup ratio for different inference models on one type of GPU (V100). The inference models (see appendix for more details) include OD (object detection), GAN (CycleGAN \cite{zhu2017unpaired}), TC (text classification) and IC (image classification). The evaluation shows a wide range of speedup ratios, from 3.6x to 47.4x. We use the model latency with CPU as each service's SLO and input them to our system. The system can recommend the best batch size and the speedup ratio of each service as a reference.

\begin{figure}[h]
\begin{subfigure}{0.5\columnwidth}
  \centering
  \includegraphics[width=1.0\linewidth]{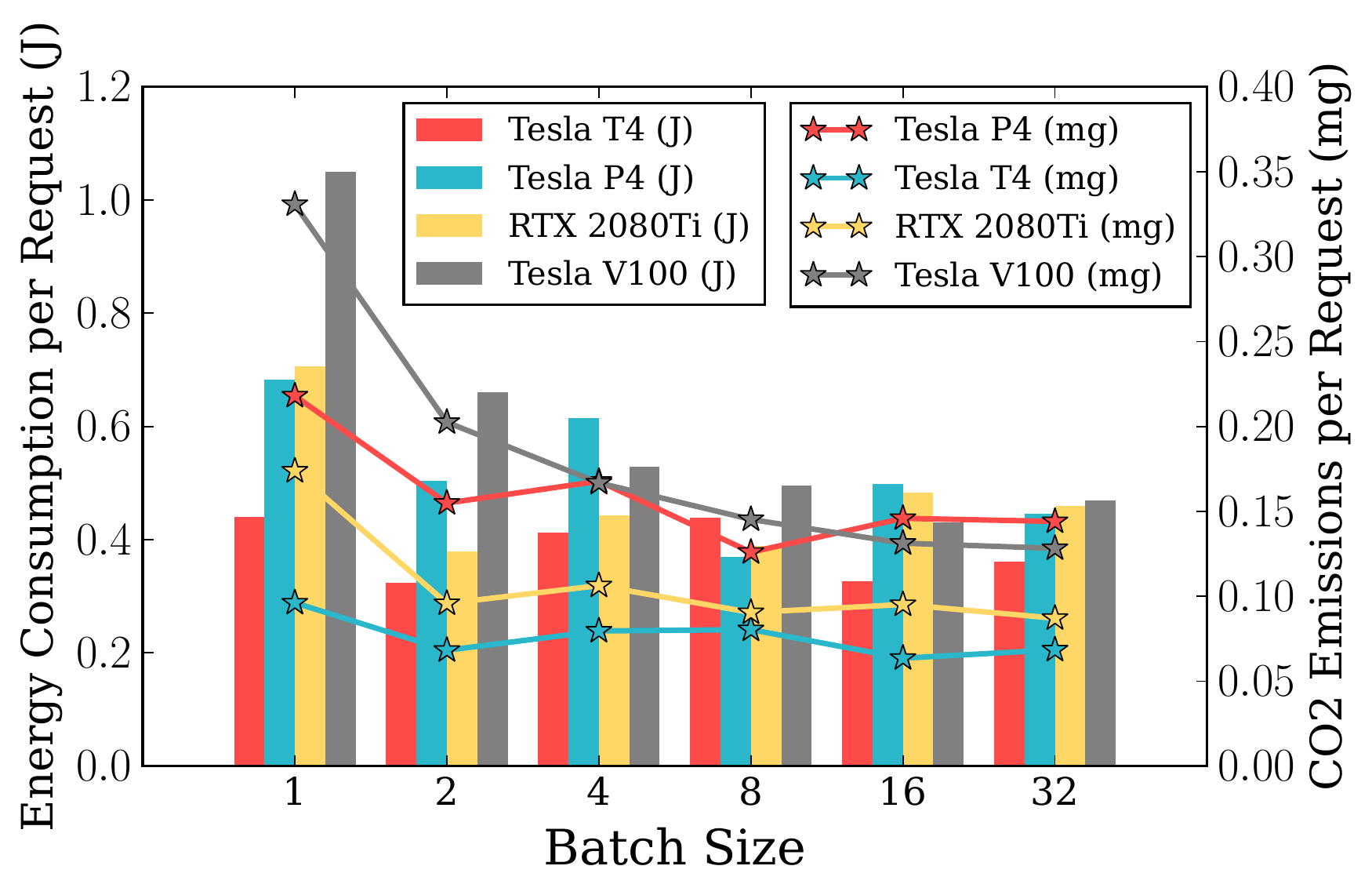}
  \caption{ResNet50 engergy and CO2 emission.}
  \label{fig:co2_energy}
\end{subfigure}%
\begin{subfigure}{.5\columnwidth}
  \centering
  \includegraphics[width=1.0\linewidth]{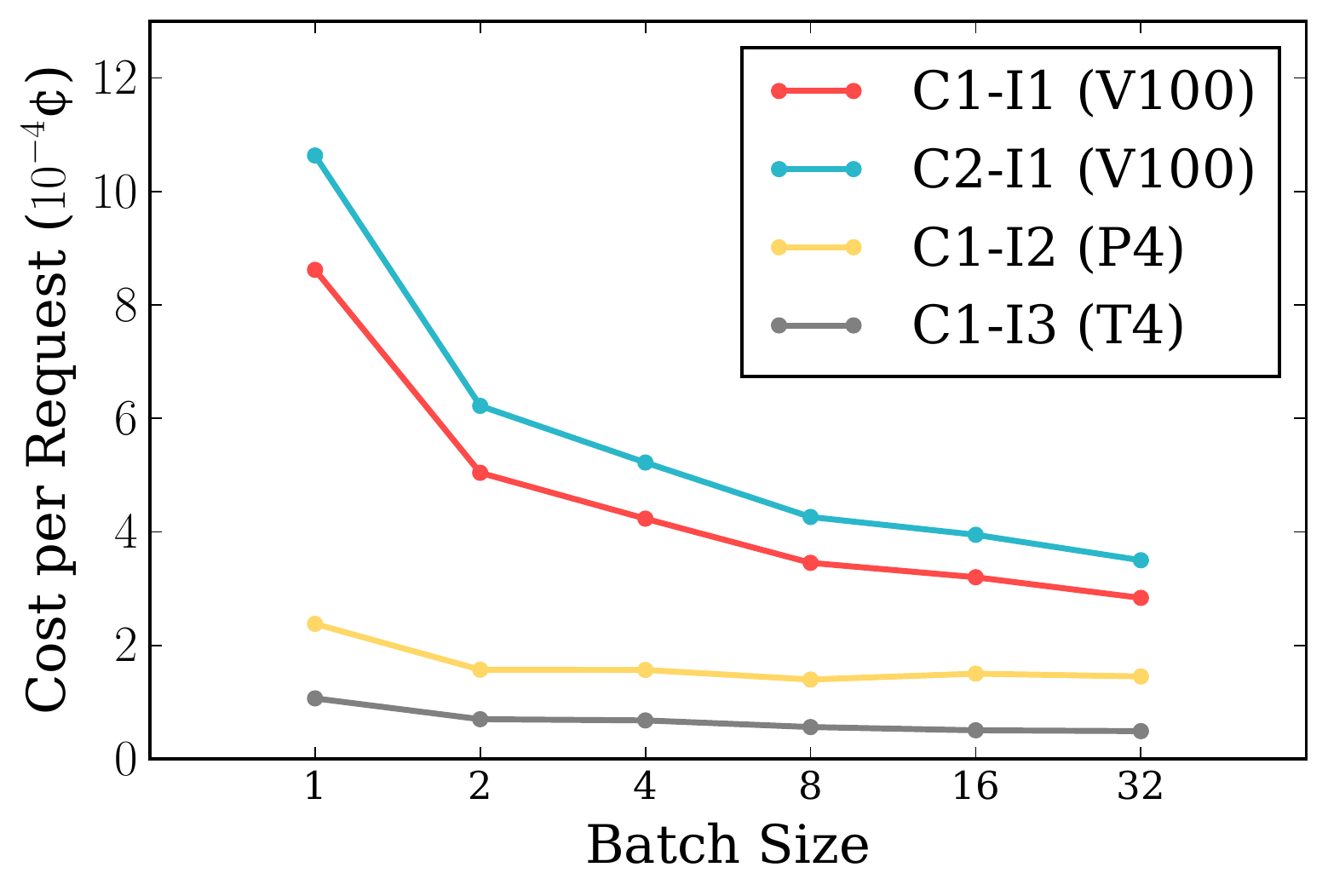}
  \caption{ResNet50 cloud cost.}
  \label{fig:cloud_cost}
\end{subfigure}
\caption{The three cost comparison across different GPUs.}

\label{fig:hardware_cost}
\end{figure}

\textbf{Cost.} We examine three kinds of costs, including energy consumption, CO2 emission, and cloud cost. Figure \ref{fig:co2_energy} shows the energy consumption and CO2 emission per request of the ResNet50 model in a batch-processing manner. In this scenario, more powerful GPUs consume more energy and emit more CO2. We note that most energy is consumed with the batch size one. This can be attributed to the overhead associated with context start, which can be amortized with larger batch sizes.

For cloud cost, we use hourly rates of different GPU instances from Google Cloud Platform and AWS. Our focus is the benchmark comparison provided by our system instead of ranking providers, so we use [C1, C2] and [I1, I2, I3] as labels for providers and instances, respectively. From the results plotted in Figure \ref{fig:cloud_cost}, we observe that 1) for the same devices (V100), different providers have different hourly rates; 2) GPU devices' costs vary with computational capability. Though T4 (I3) GPU is more powerful than P4 (I2) GPU, it has a lower price; and 3) as the batch size increases, more images can be processed hourly. As a result, the cost per request decreases. With this capability provided by our system, users can choose the best cloud providers and instances for their services.

\begin{figure}[t]
\begin{subfigure}{0.5\columnwidth}
  \centering
  \includegraphics[width=1.0\linewidth]{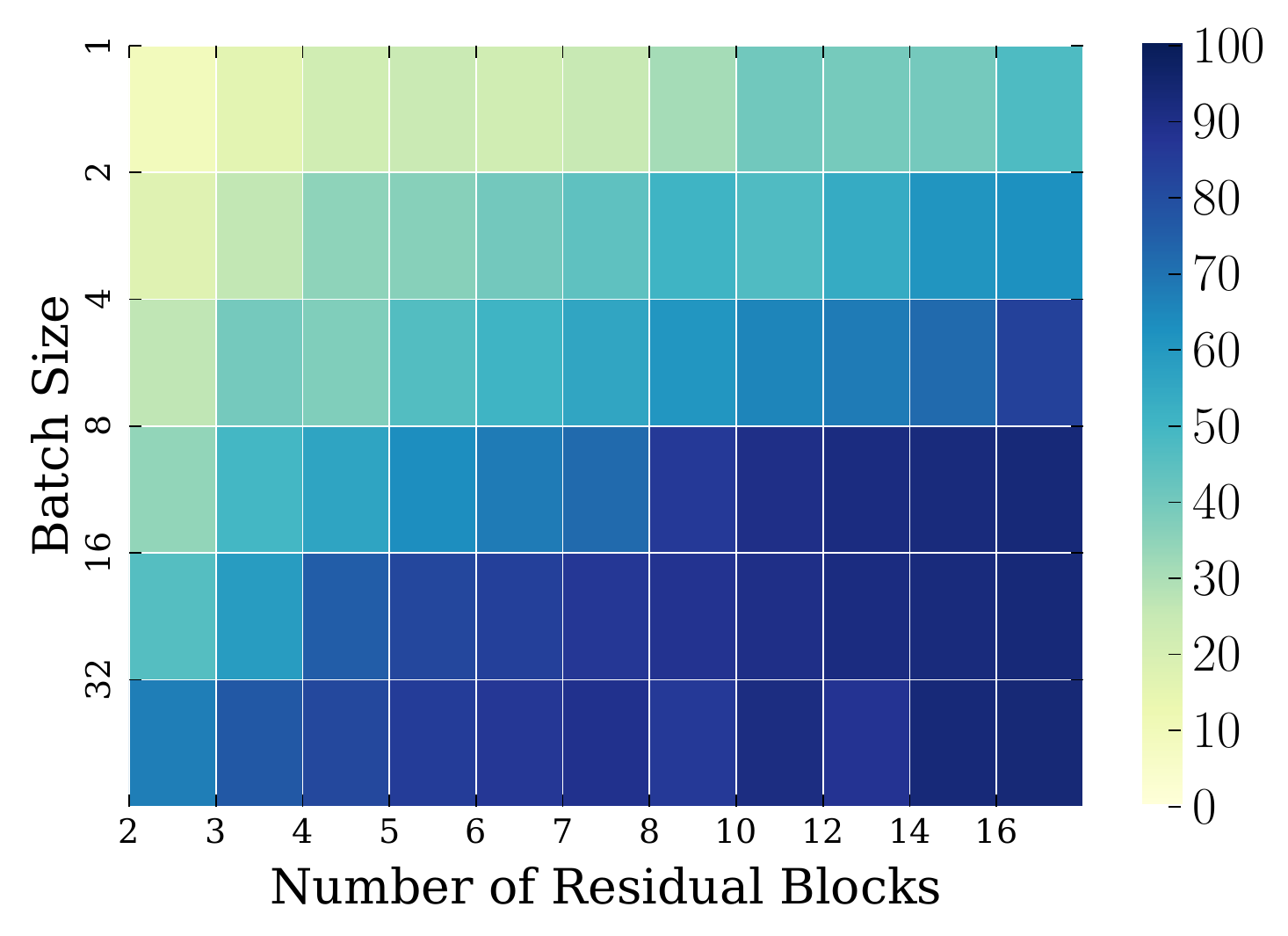}
  \caption{CNN models.}
  \label{fig:heatmap_cnn}
\end{subfigure}%
\begin{subfigure}{0.5\columnwidth}
  \centering
  \includegraphics[width=1.0\linewidth]{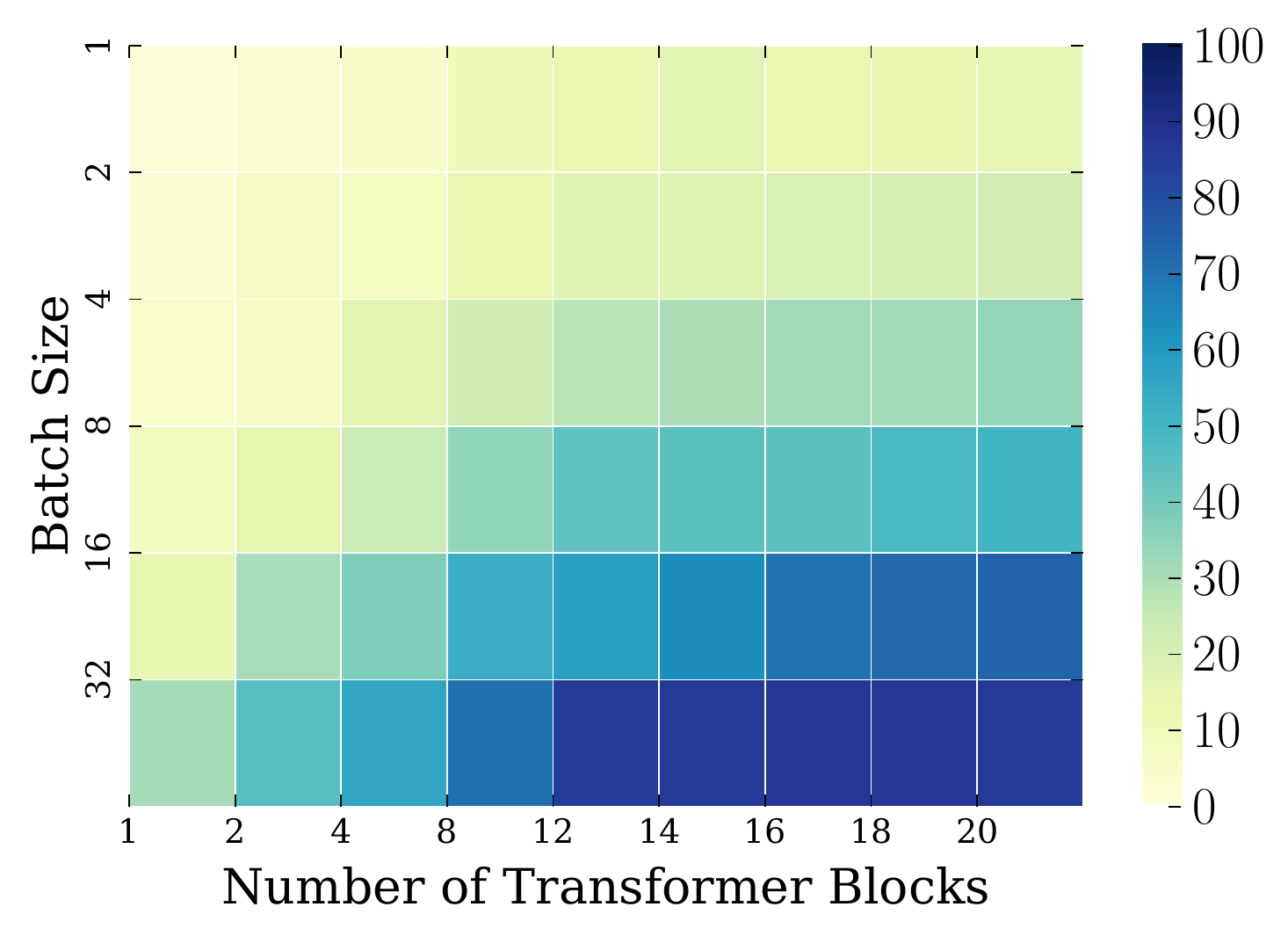}
  \caption{Transformer models.}
  \label{fig:heatmap_transformer}
\end{subfigure}
\caption{The GPU utilization with varied hyper-parameters of different models.}

\label{fig:hardware_heatmap}
\end{figure}

\textbf{Performance Sensitivity.} We evaluate performance sensitivity to hyper-parameters such as batch size and the layer number with generated models on a V100 GPU. Every time we select two parameters and keep the others fixed. Figure \ref{fig:hardware_heatmap} shows two examples. For a CNN type model, GPU utilization increases with both batch size and model depth, indicating that a GPU exploits parallelism within the batch size and depth of this kind of models. For a transformer model, the model's depth has more impact, indicating that a transformer model with more layers will utilize GPU resources more. Since more hardware accelerators (e.g., TPU) are emerging, we provide a powerful tool for hardware engineers to explore the hyper-parameter influences.

\begin{figure}[h]
\begin{subfigure}{0.5\columnwidth}
  \centering
  \includegraphics[width=1.0\linewidth]{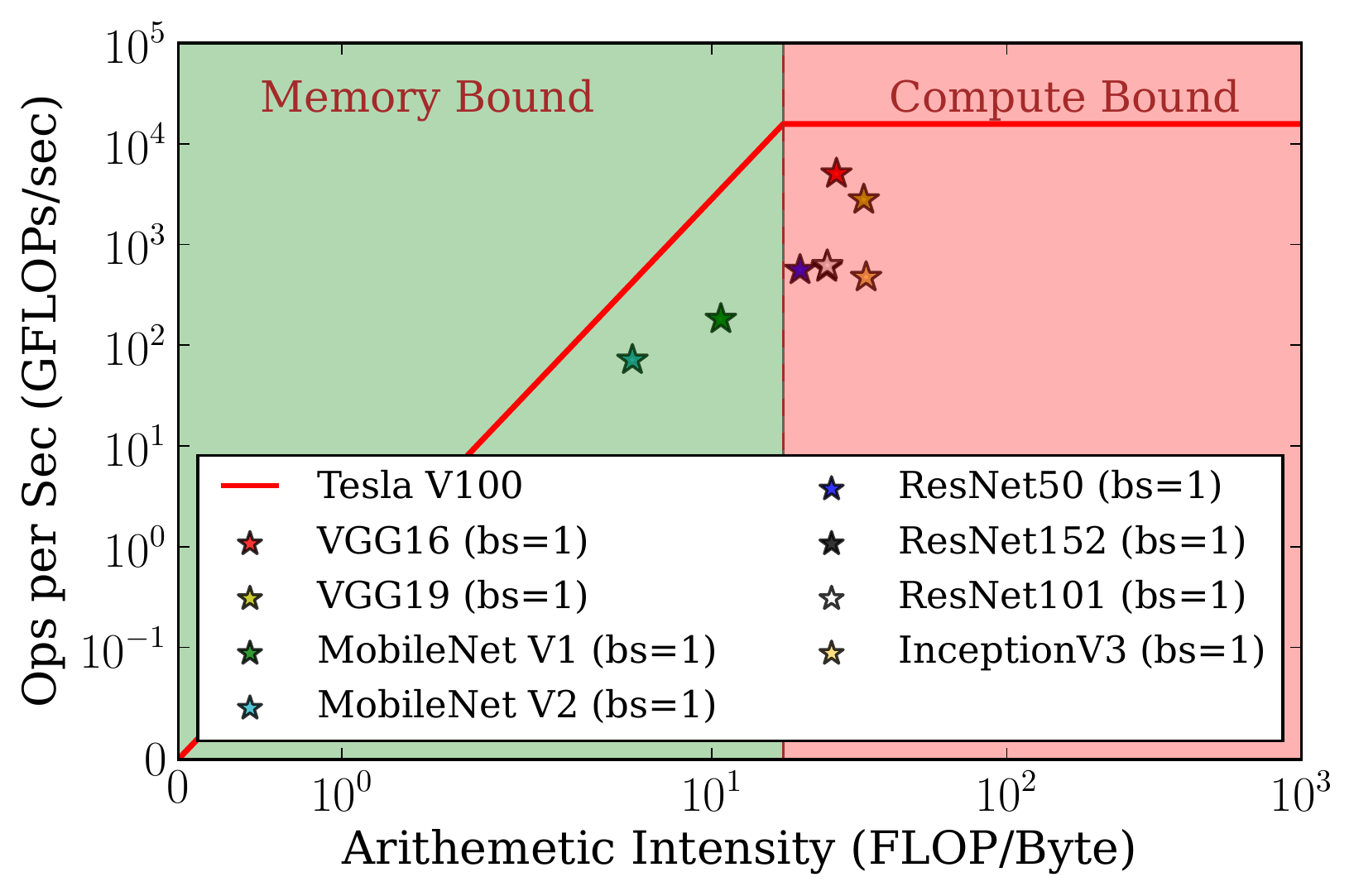}
  \caption{Real-world models}
  \label{fig:roofline_real}
\end{subfigure}%
\begin{subfigure}{0.5\columnwidth}
  \centering
  \includegraphics[width=1.0\linewidth]{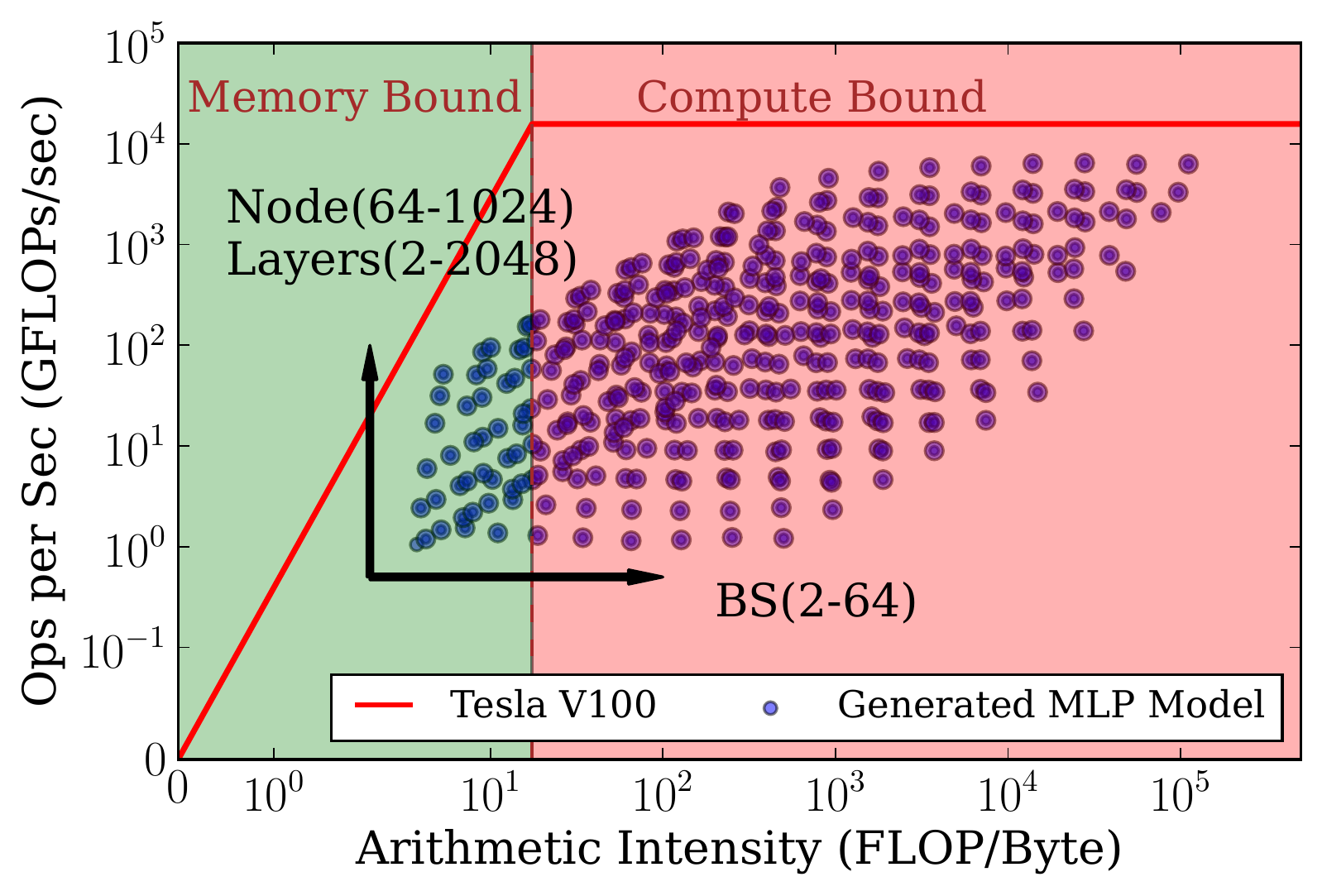}
  \caption{Generated models}
  \label{fig:roofline_fc}
\end{subfigure}
\caption{The Roofline model analysis with both real-world models and generated models.}

\label{fig:hardware_roofline}
\end{figure}

\textbf{Computation and Memory Bandwidth}. The performance of a GPU is decided by both its computation capacity and memory bandwidth. We apply Roofline models, which represent the relationship between the model's operational intensity and operations per second, to give a quantitative comparison. The ceiling line (in red) shows the theoretical bandwidth and computation capability of a GPU (V100). For real-world CNN models (shown in Figure \ref{fig:roofline_real}), we observe that two lightweight models (MobileNet \cite{howard2017mobilenets}) are more memory-bound and the other heavy models are more compute-bound. This is in line with that in real-world practice, the performance improvement of MobileNet does not significantly align with its small numbers of parameters. It can not fully utilize the computation of GPUs. With our system, data scientists can easily understand their models for optimization. Figure \ref{fig:roofline_fc} presents the Roofline analysis of generated models. The model operations per second increases while the arithmetic intensity increases. Larger batch sizes make MLP models more compute-bound, whereas more layers and more neurons incur a memory-bound. This insight can not be obtained from isolated real-world models. Hardware engineers can apply our system to analyze the inference performance of both newly proposed model structures or accelerators.

\subsection{Software Platform Characterization}
\label{sec:software_exp}

We prepare several benchmark submissions to evaluate software platforms. They serve models with a range of up to 5 minutes. The workload generator is used to simulate Poisson Distribution with varied arrival rates to understand software performance. We also explore the advanced features like the dynamic batching specific to DL serving.

\begin{figure}[t]
\begin{subfigure}{0.5\columnwidth}
  \centering
  \includegraphics[width=1.0\linewidth]{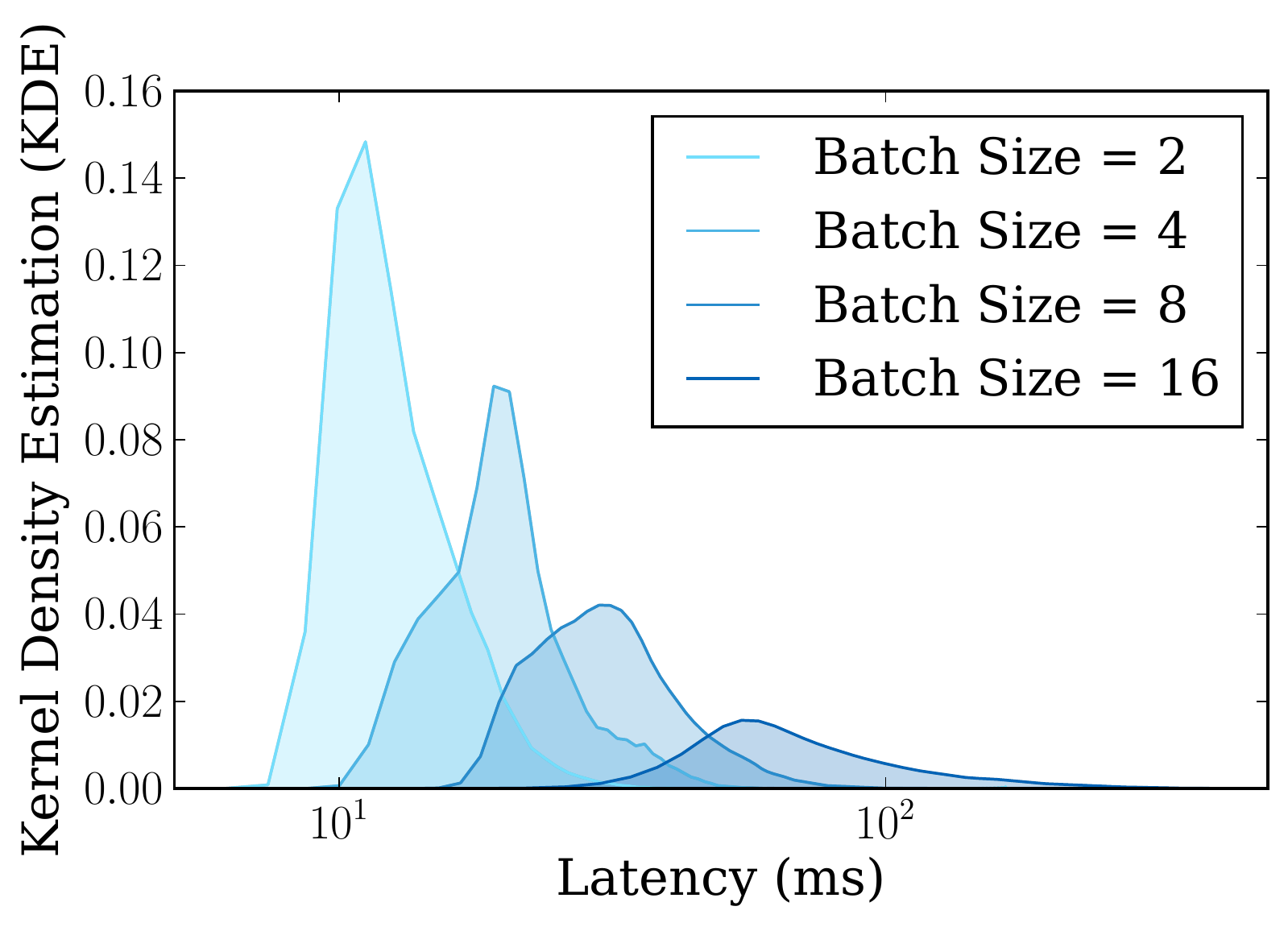}
  \caption{Batch size}
  \label{fig:kde_batch_size}
\end{subfigure}%
\begin{subfigure}{.5\columnwidth}
  \centering
  \includegraphics[width=1.0\linewidth]{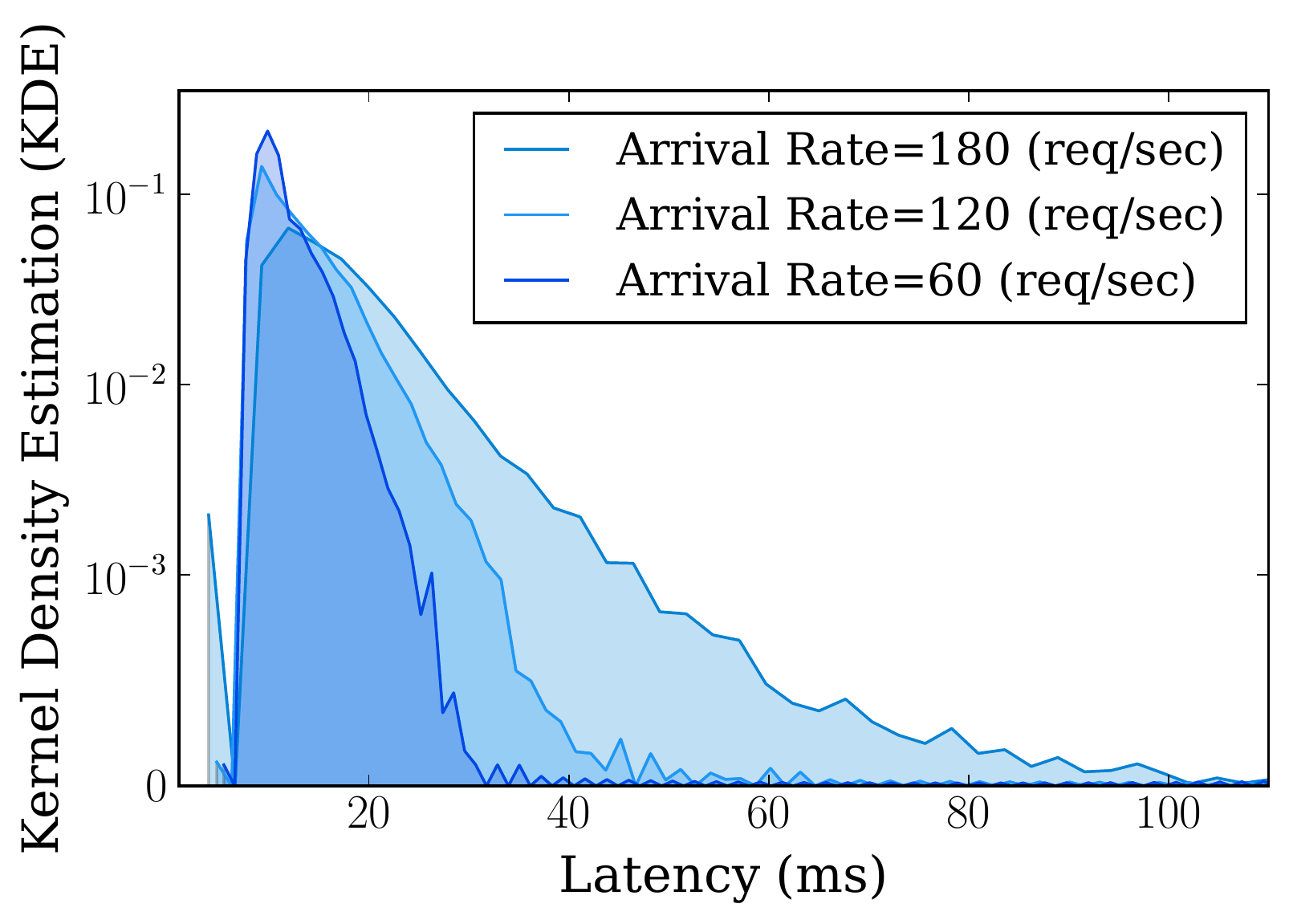}
  \caption{Arrival rate (KDE)}
  \label{fig:kde_tfs_arrival_rate}
\end{subfigure}
\begin{subfigure}{0.5\columnwidth}
  \centering
  \includegraphics[width=1.0\linewidth]{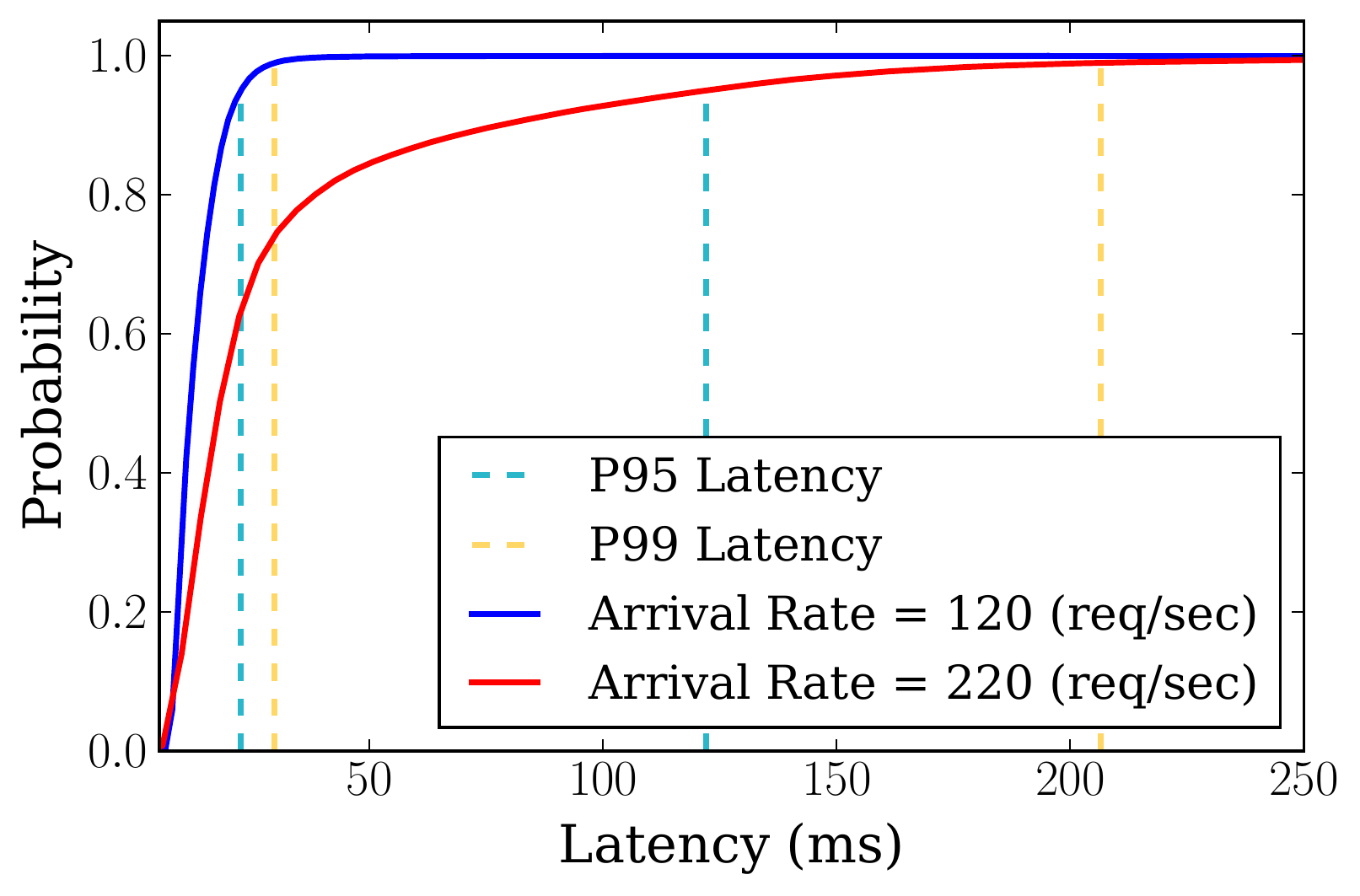}
  \caption{Arrival rate(CDF)}
  \label{fig:cdf_arrival_rate_tfs}
\end{subfigure}%
\begin{subfigure}{.5\columnwidth}
  \centering
  \includegraphics[width=1.0\linewidth]{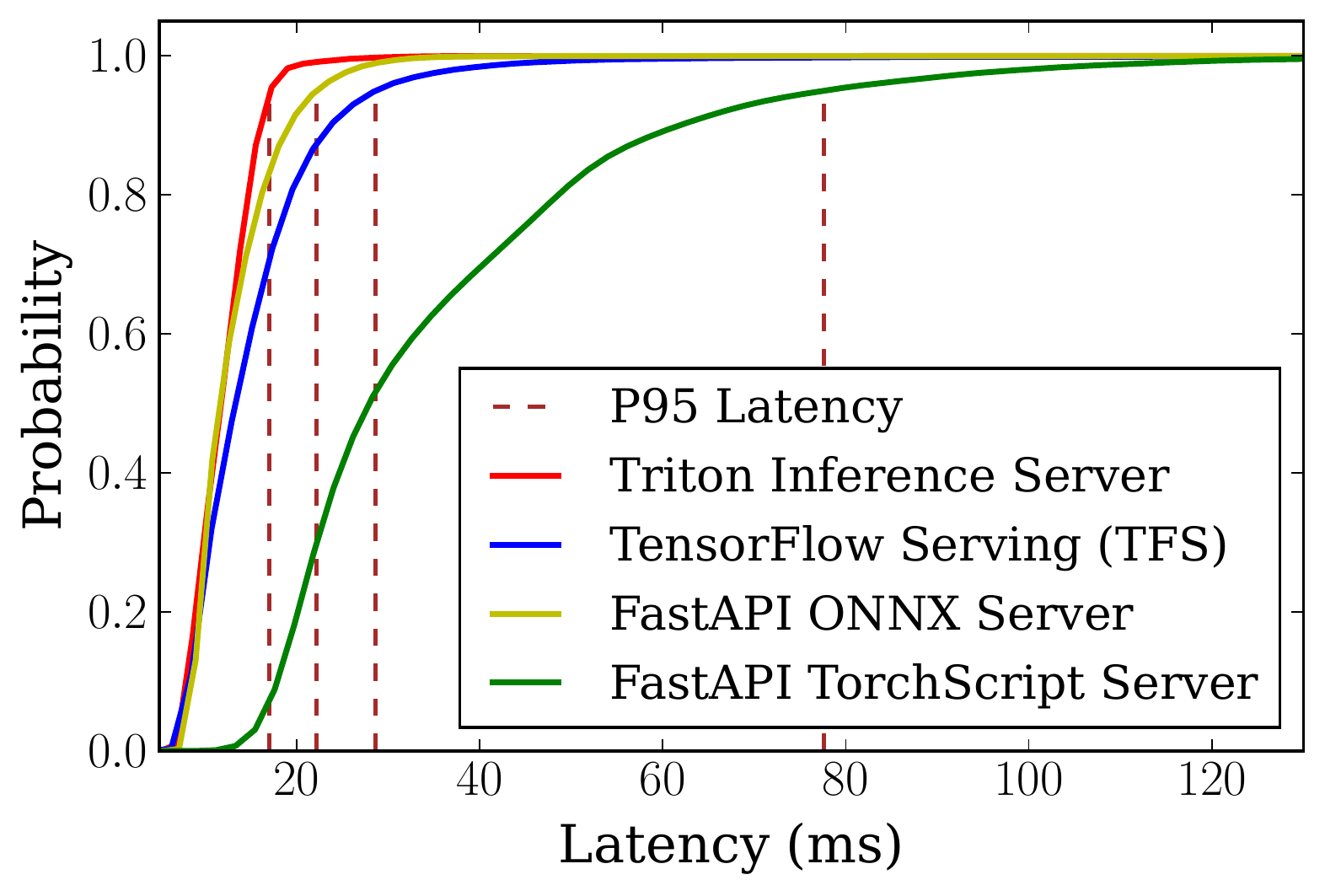}
  \caption{Four software}
  \label{fig:cdf_arrival_rate_four_platform}
\end{subfigure}
\caption{The tail latency of four systems under varied workloads. The batch size, arrival rate, and serving software influence tail latency a lot.}
\label{fig:software_tail_latency}

\end{figure}

\textbf{Tail Latency.} We compare the tail latency across varied software and arrival rates. We use TFS with ResNet50 as a case study. As shown in Figure \ref{fig:kde_batch_size}, the larger batch size accounts for a longer tail latency though it can save service providers' cost. Both Figure \ref{fig:kde_tfs_arrival_rate} and Figure \ref{fig:cdf_arrival_rate_tfs} indicate that TFS can not adequately handle spike load.

We further compare different four platforms for a same image classification service (based on ResNet50) on a V100 GPU as shown in Figure \ref{fig:cdf_arrival_rate_four_platform}. TrIS from NVIDIA performs best with no surprise since it includes many GPU optimization techniques. ONNX Runtime with a simple web framework performs better than TFS, a dedicated serving system for DL inference. In conclusion,  when configuring a latency-intensive service, developers need to take the impact of batch size and the served software (on latency) into consideration. Both academia and industry have invested efforts in serving software development, and our system can make things easier.

\textbf{Advance Feature: Dynamic Batching.} Two representative systems, TFS and TrIS, provide dynamic batching settings. To use the feature, developers need to set a lot of hyper-parameters like the maximum batch size and maximum queue time. We use the two software to serve a ResNet50 model and send requests concurrently. In this case, TrIS can utilize the feature and improve the throughput steadily while TFS performs even worse than no dynamic batching in a small concurrency. This reminds us that before starting to use the feature, engineers should understand their scenarios first and tune the two parameters accordingly. 

\begin{figure}[h]
\begin{subfigure}{0.5\columnwidth}
  \centering
  \includegraphics[width=1.0\linewidth]{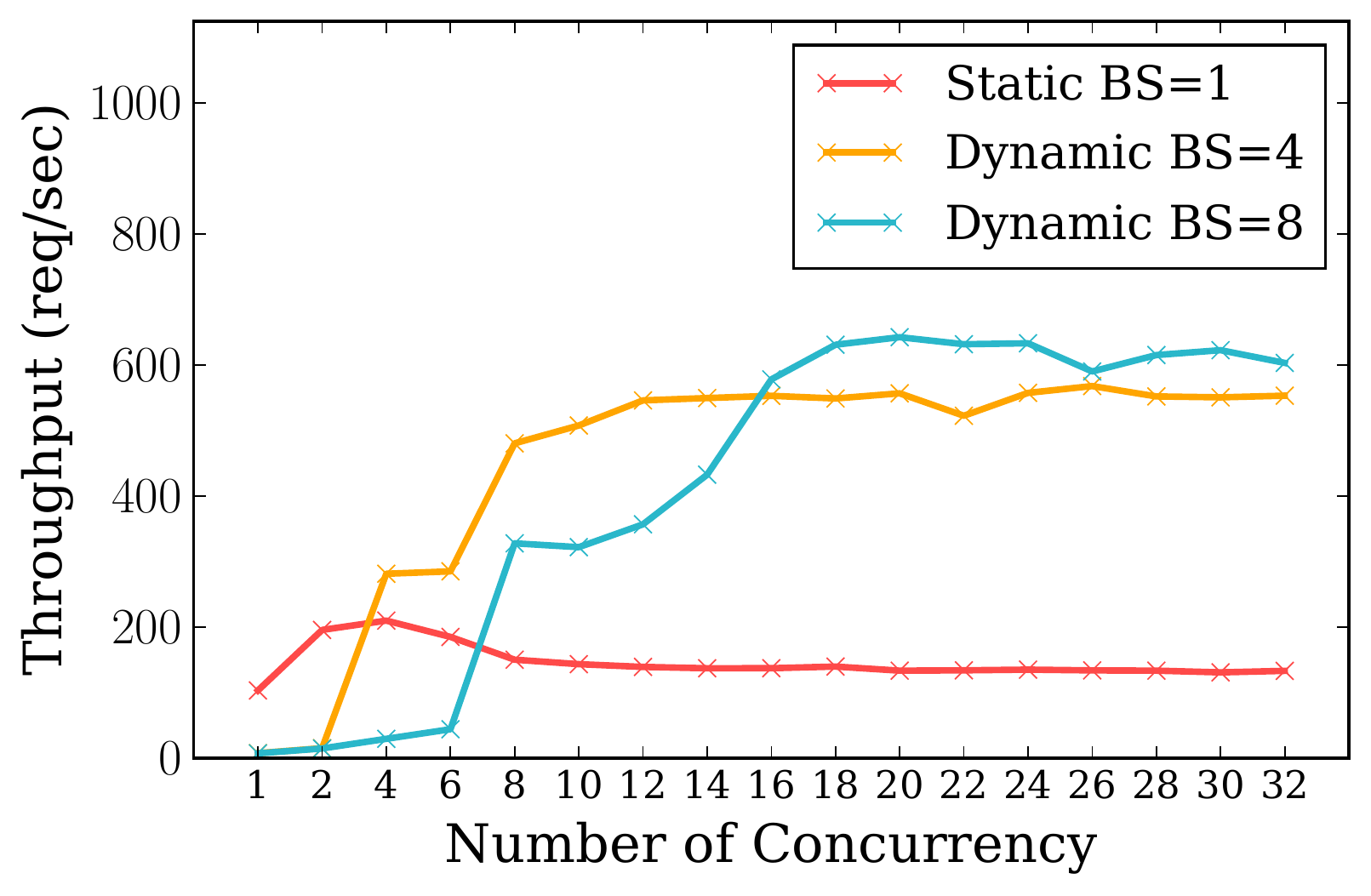}
  \caption{ResNet50 served by TFS}
  \label{fig:software_tfs_dynamic_batching}
\end{subfigure}%
\begin{subfigure}{.5\columnwidth}
  \centering
  \includegraphics[width=1.0\linewidth]{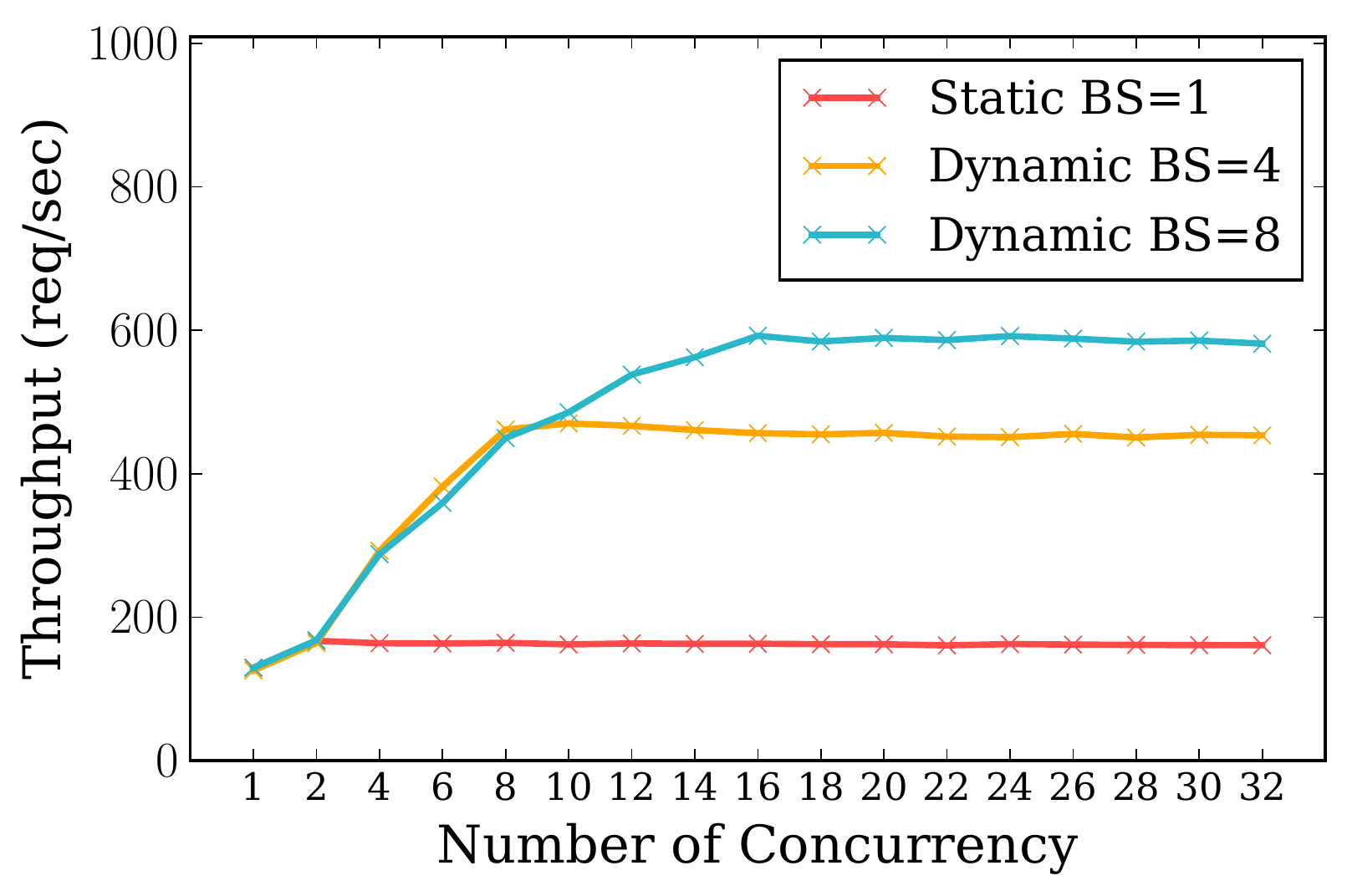}
  \caption{ResNet50 served by TrIS}
  \label{fig:software_trts_dynamic_batching}
\end{subfigure}
\caption{The throughput comparison of two serving software with the dynamic batching feature.}

\label{fig:software_dynamic_batching}
\end{figure}

\begin{figure}[ht]
\begin{subfigure}{0.5\columnwidth}
  \centering
  \includegraphics[width=1.0\linewidth]{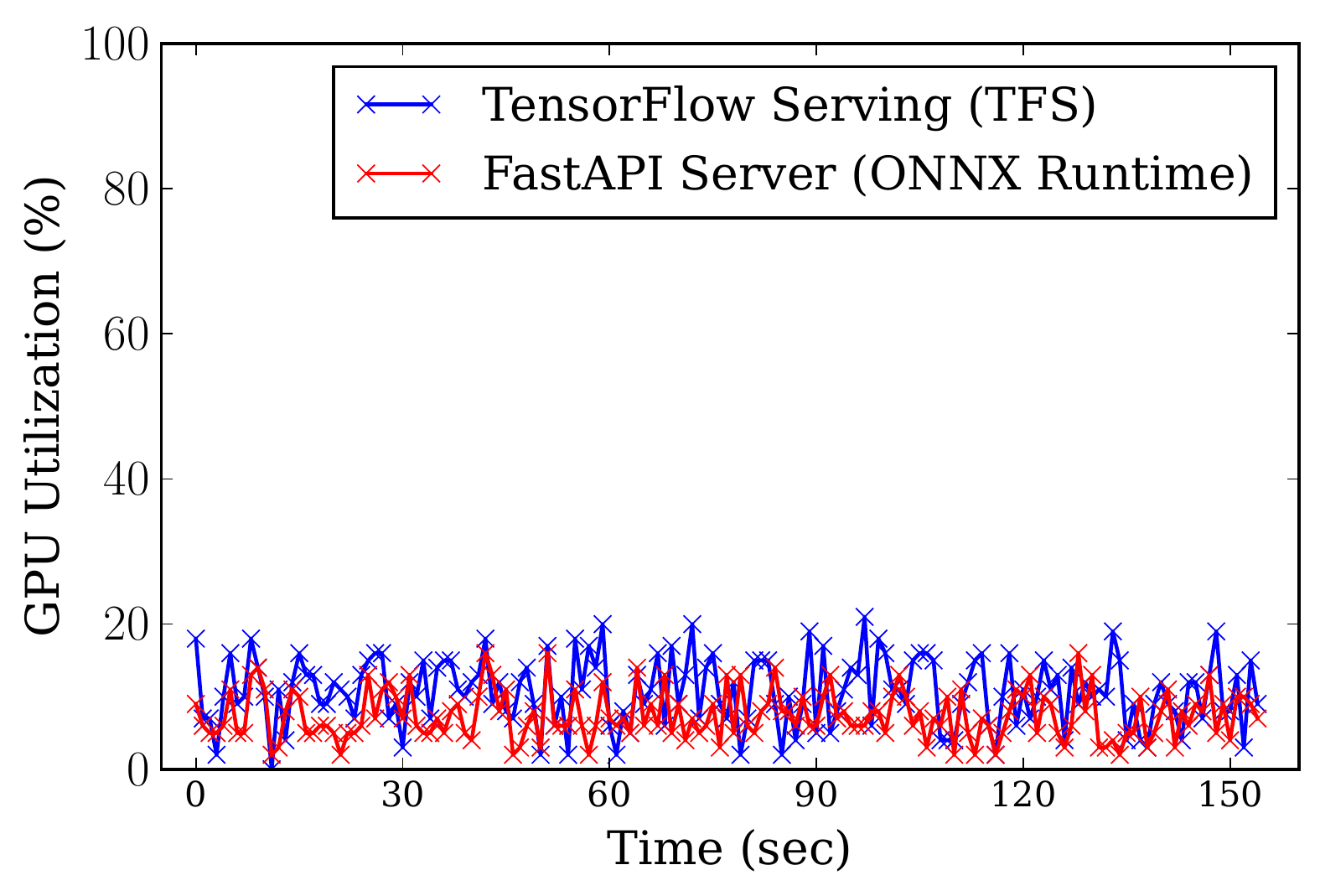}
  \caption{BERT}
  \label{fig:software_utilization_resnet50}
\end{subfigure}%
\begin{subfigure}{.5\columnwidth}
  \centering
  \includegraphics[width=1.0\linewidth]{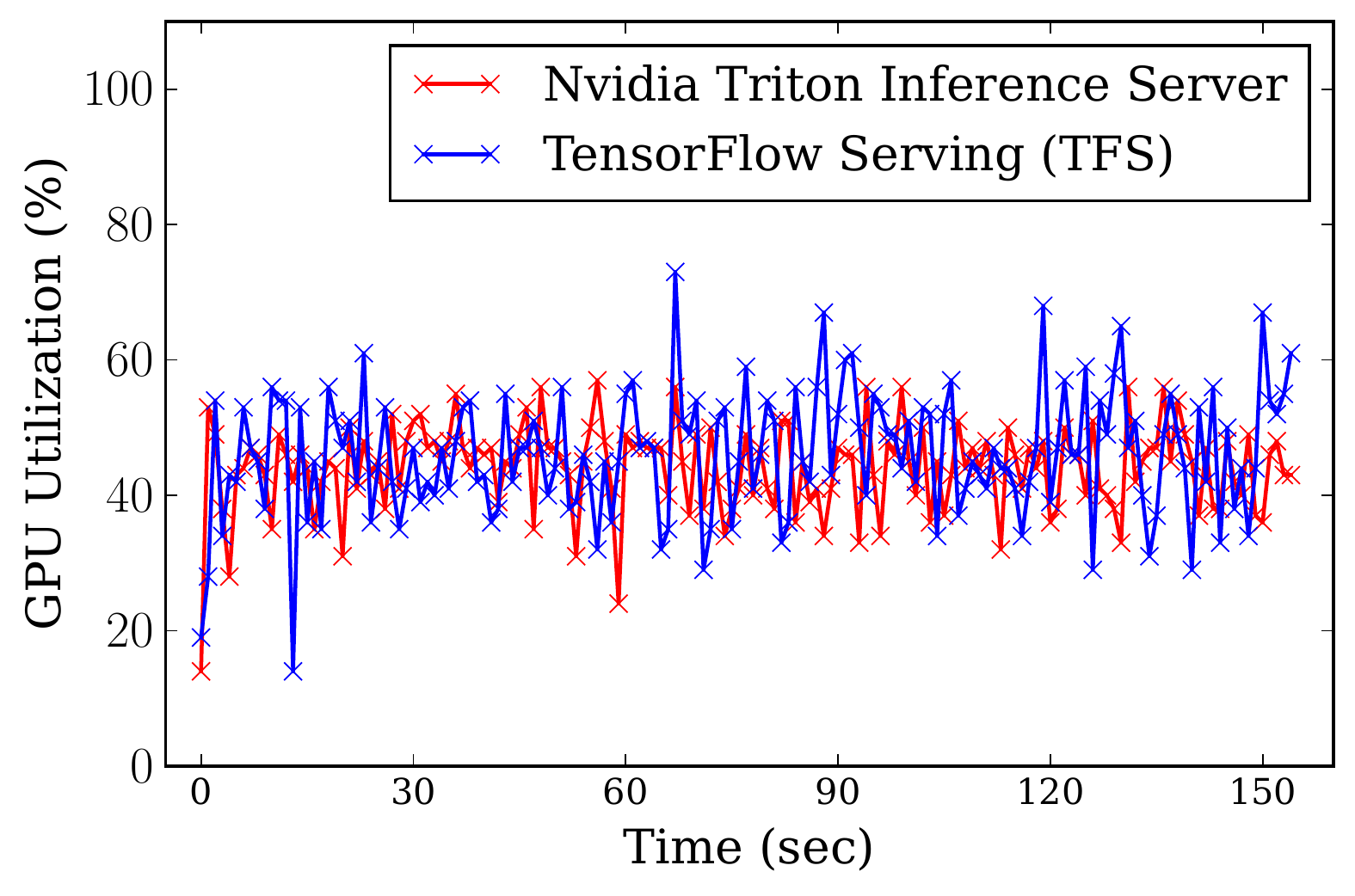}
  \caption{ResNet50}
  \label{fig:software_utilization_bert}
\end{subfigure}
\caption{The GPU utilization with different serving software under varied workloads}
\label{fig:software_gpu_utilization}
\end{figure}

\textbf{Resource Usage.} Understanding the resource utilization pattern often leads to better resource allocations. We test two services (BERT with an arrival rate of 30 requests/seconds and batch size 1 and ResNet50 with 160 requests/second and batch size 1). The results are shown in Figure \ref{fig:software_gpu_utilization}. We observe that the GPU utilization is dynamic with varied workloads and tends to be under-utilization with a low arrival rate (even it loads a heavy model like BERT). This gives developers a large enough room for optimization.

\subsection{Inference Pipeline Decomposition}

We simulate a real-world DL service by building a simple pipeline and sending requests using three networking transmission technologies. We use the ResNet50 and TFS as the inference model and the serving system, respectively.

\begin{figure}[t]
\begin{subfigure}{0.5\columnwidth}
  \centering
  \includegraphics[width=1.0\linewidth]{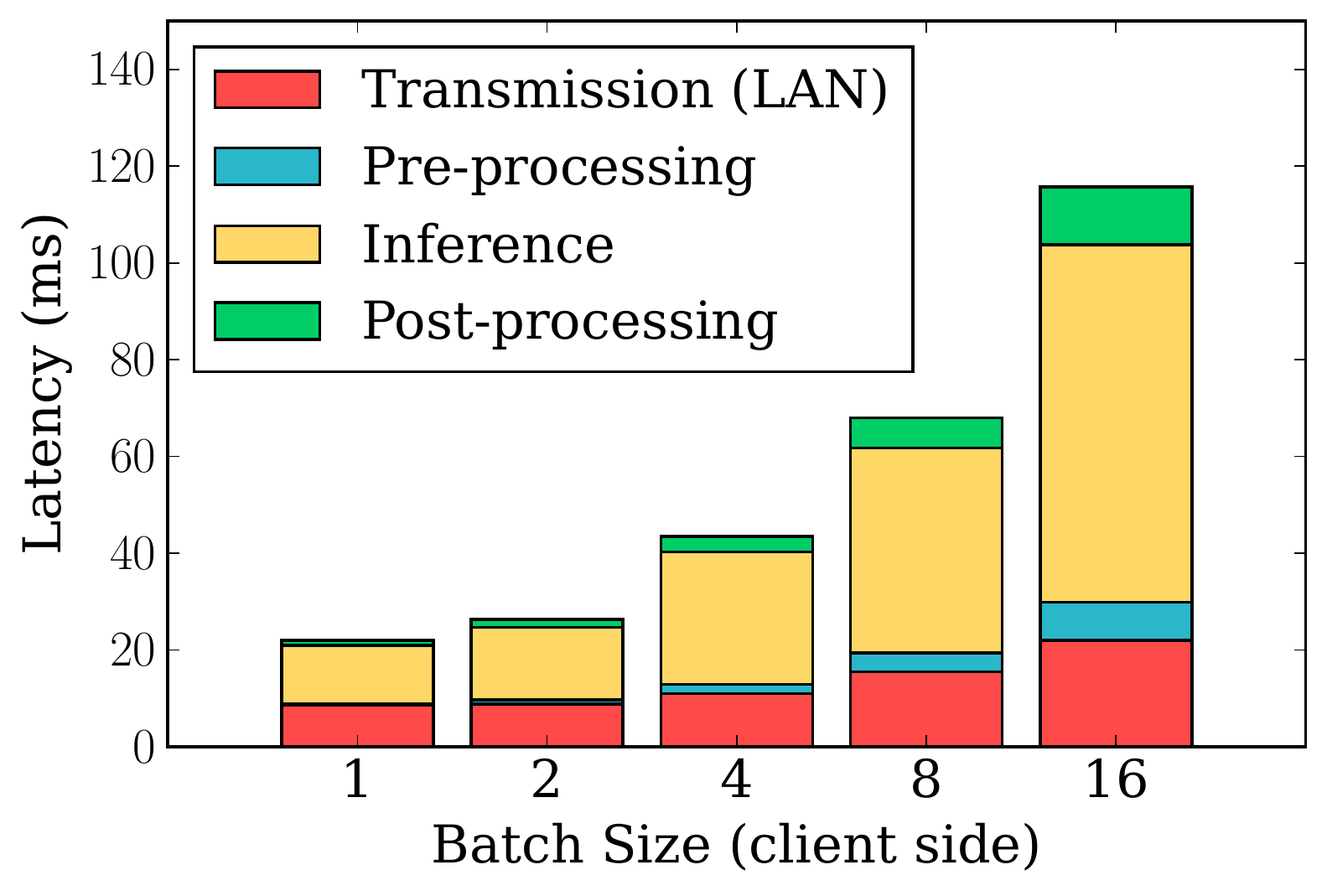}
  \caption{Batch size}
  \label{fig:pipeline_LAN}
\end{subfigure}%
\begin{subfigure}{.5\columnwidth}
  \centering
  \includegraphics[width=1.0\linewidth]{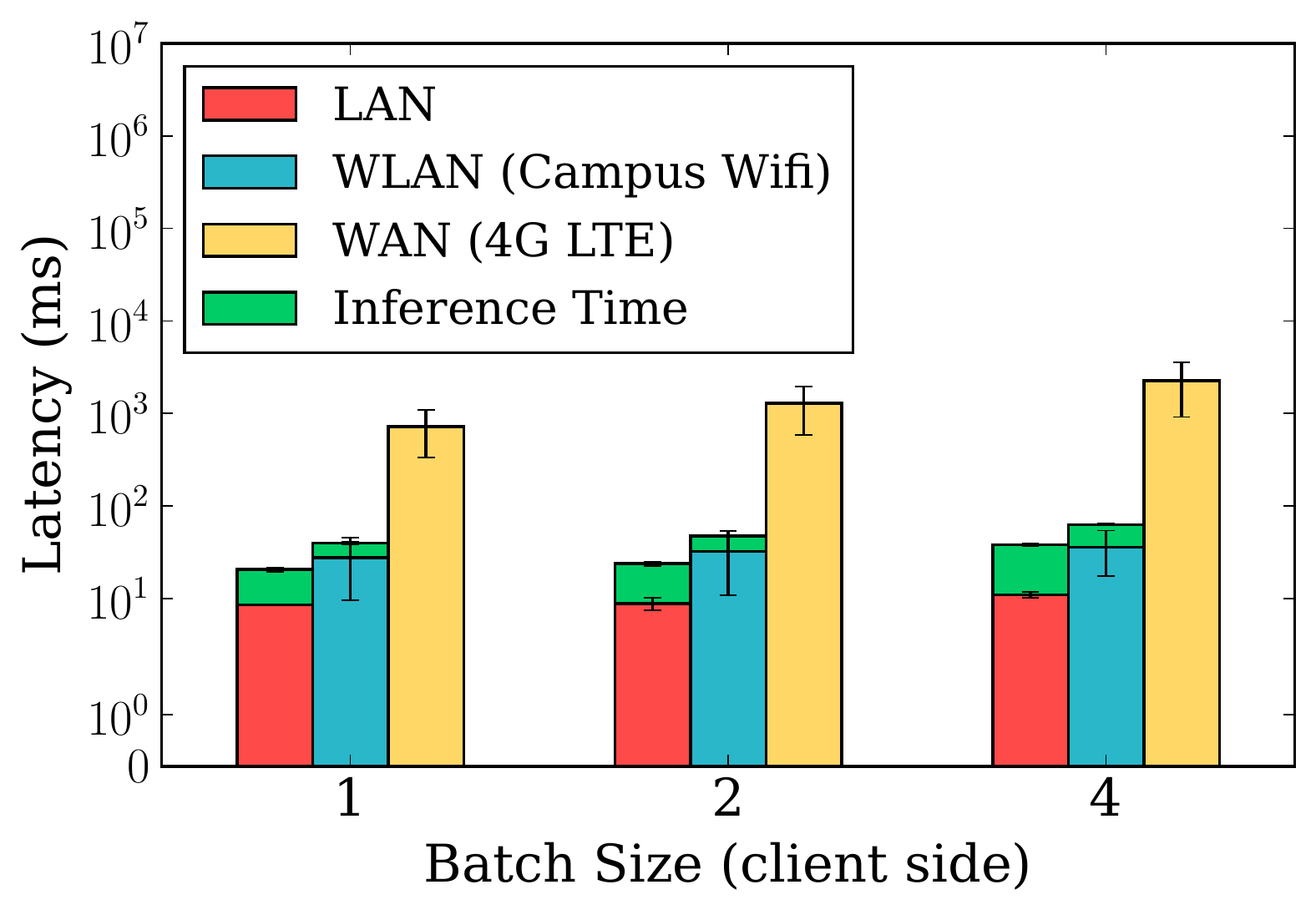}
  \caption{Networking condition}
  \label{fig:pipeline_others}
\end{subfigure}
\begin{subfigure}{1.0\columnwidth}
  \centering
  \includegraphics[width=1.0\linewidth]{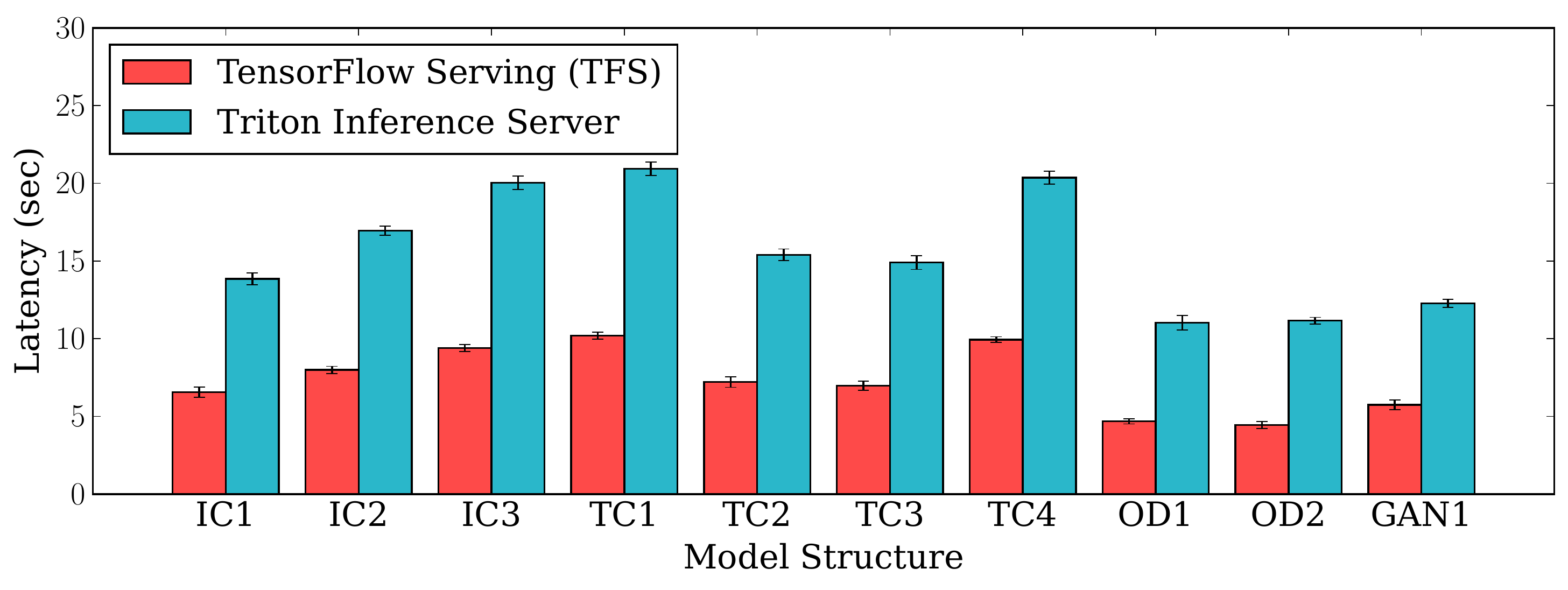}
  \caption{The cold start latency of different models with two software}
  \label{fig:cold_start}
\end{subfigure}
\caption{The pipeline decomposition results.}

\label{fig:pipeline_decomp}
\end{figure}

The results in Figure \ref{fig:pipeline_LAN} show that the transmission time is very comparable to inference time for small batch sizes. As the batch size increases, the inference time accounts for a much larger portion of the total time. The results (Figure \ref{fig:pipeline_others}) also show that for the same service, 4G LTE has the longest end-to-end latency. This indicates that for a mobile application, sending requests to cloud DL service can incur high latency.

We also compare the cold start time of different models with two software. TrIS has a longer starting time than that of TFS. Even for a small image classification model, it needs more than 10 seconds to prepare. In practice, the long starting time can pose challenges for resource provisioning.

\subsection{Case Study: Task Scheduling}

\begin{figure}[ht]
\begin{subfigure}{0.5\columnwidth}
  \centering
  \includegraphics[width=1.0\linewidth]{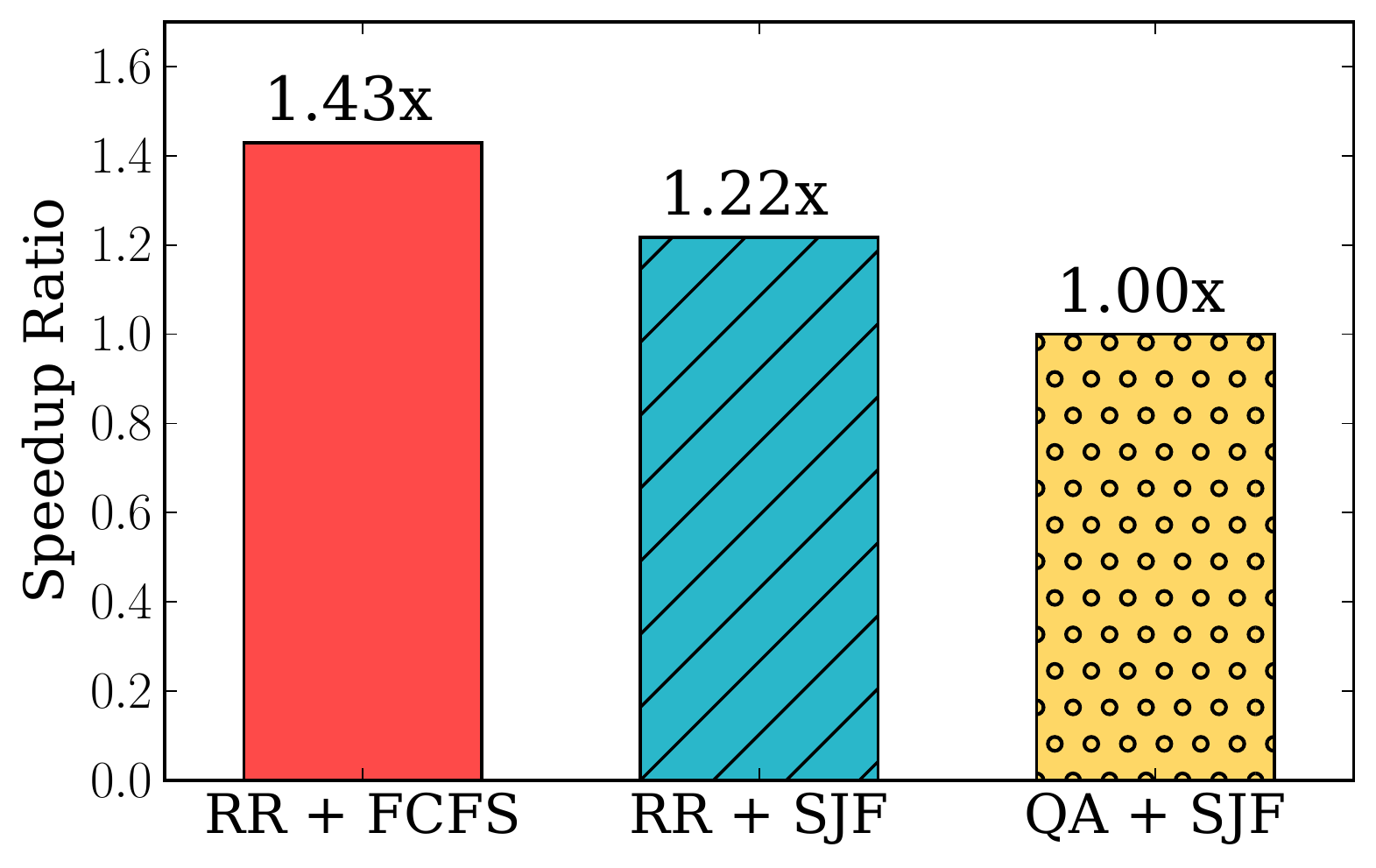}
  \caption{Speedup ratio}
  \label{fig:errobar_jct}
\end{subfigure}%
\begin{subfigure}{.5\columnwidth}
  \centering
  \includegraphics[width=1.0\linewidth]{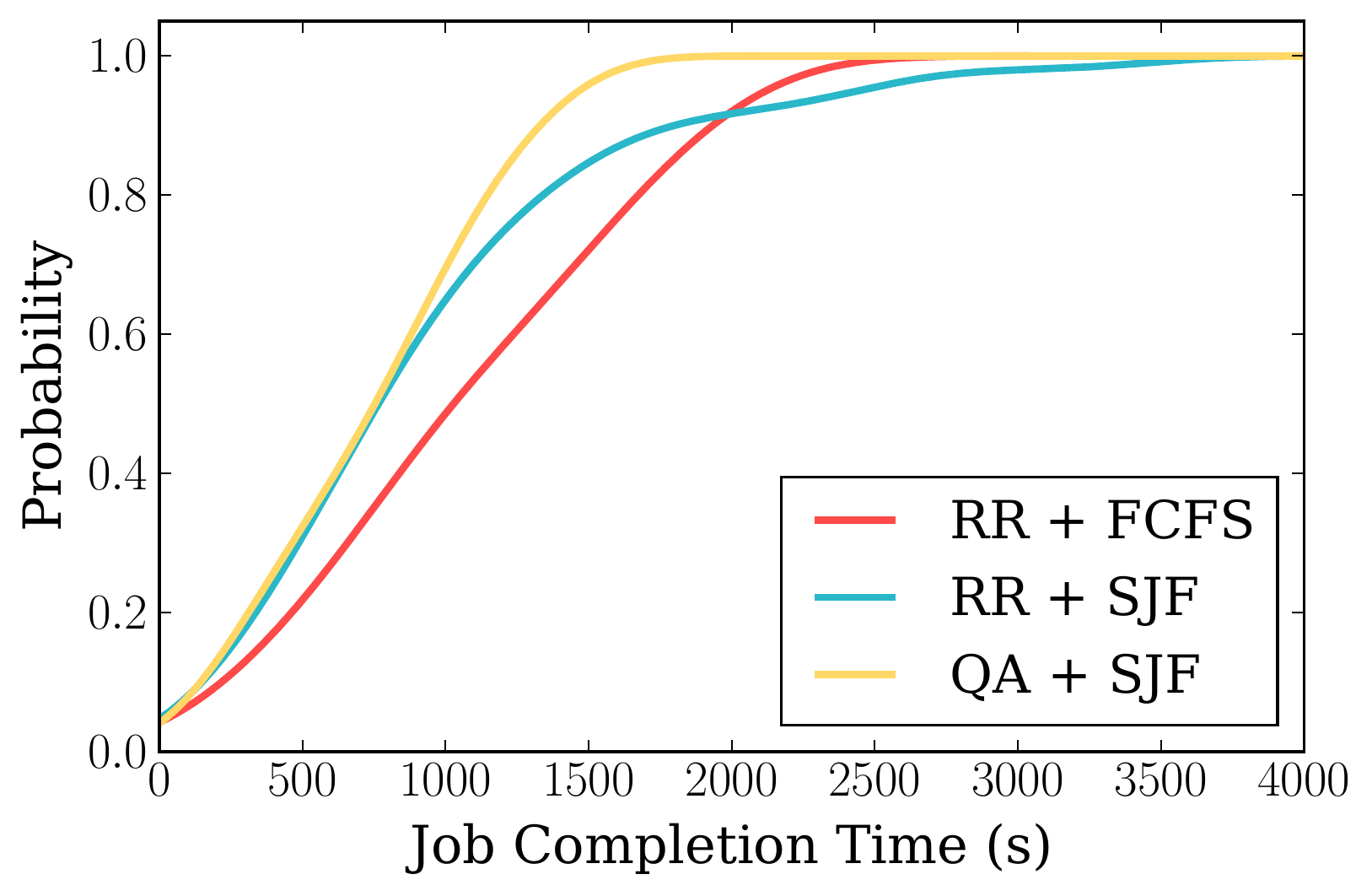}
  \caption{CDF}
  \label{fig:cdf_jct}
\end{subfigure}
\caption{The performance comparison of three schedulers.}

\label{fig:jct}
\end{figure}

To the best of our knowledge, current benchmarking tasks are often carried out in an error-to-prone fashion. Since tasks need to be executed in an idle server, the system status needs to be checked. If this check is neglected, all tasks located in a worker will crash. As there are no specific scheduling methods to address these issues before this work, we implement two methods as the baselines, a round-robin (RR) load balancer (LB) with First-Come-First-Serve (FCFS) and an LB with Short-Job-First (SJF). The results \ref{fig:jct} show that our scheduler, queue aware (QA) LB with SJF, can reduce the average job-completion-time JCT by 1.43 (equivalent of 30\% reduction). Now we assume that the processing time of every benchmark task is determined before they are executed. We leave the study of a scheduler for jobs with stochastic processing time for future work.

\section{Related Work}
\label{sec:relatedwork}

The system's design follows the best practice of previous benchmarking works and we provide a highly efficient cluster system. We categorize previous studies into two classes: micro-benchmark and macro-benchmark. Micro-benchmark focuses more on low-level operators. DeepBench \cite{deepbench} is designed to study the kernel-level operations such as GEMM across multiple hardware devices, including both cloud servers and edge nodes. AI matrix \cite{aimatrix} gets inspiration from DeepBench and further extends it to benchmark both layers (e.g., RNN) and models (e.g., FasterRCNN). These studies focus on small sets of impact factors and can not simulate the scenarios in practice. Macro-benchmark such as Fathom \cite{adolf2016fathom} and AI Benchmark \cite{ignatov2019ai} collect a set of models and study their performance on the mobile side. DawnBench \cite{coleman2017dawnbench} and MLPerf Inference \cite{reddi2020mlperf} support more scenarios and provide competition for both industry and academia to measure their systems. In comparison, our work provides full support for developers to speed up their benchmark process and configure new services with ease. Our system also explores a wider space to evaluate DL systems for a comprehensive study. ParaDNN \cite{wang2020systematic} also uses analysis models like Roofline and Heat maps to study DL platforms, but it focuses on training workload and lack of configurability. In general, we implement the system to complement existing inference benchmark suits and obtain more comprehensive analysis results.

\section{Summary and Future Work}
\label{sec:summary}

With the rapid development of deep learning (DL) models and the related hardware and software, benchmarking for service configuration and system upgrade becomes developers' day-to-day tasks. Previous work focuses on the isolated DL model benchmark study and leaves tedious and error-prone tasks to developers. In this work, we design and implement an automatic benchmarking system to address these issues. Our system streamlines the task executions and explores a wide range of design space. It provides many analysis models to gain insights for resource allocation and service operations. A scheduler is implemented to improve efficiency. We conduct many experiments to demonstrate the functions of our system and provide many practical guidelines. We plan to continue upgrading the system and design more APIs to give users more flexibility to adopt our system into their own DL services

\newpage
\bibliography{example_paper}
\bibliographystyle{mlsys2020}

%


\end{document}